\documentclass[journal]{IEEEtran}
\usepackage{amsmath,amsfonts}
\usepackage{amsthm}
\usepackage[ruled,linesnumbered]{algorithm2e}
\usepackage{array}
\usepackage[colorlinks,urlcolor=blue,linkcolor=blue,citecolor=blue]{hyperref}
\usepackage{textcomp}
\usepackage{stfloats}
\usepackage{url}
\usepackage{verbatim}
\usepackage{graphicx}
\usepackage{cite}
\usepackage{multirow,booktabs}
\usepackage{balance}
\usepackage{makecell}
\usepackage{booktabs}
\usepackage{amssymb}
\usepackage{tcolorbox}

\ifCLASSOPTIONcompsoc
\usepackage[caption=false, font=normalsize, labelfont=sf, textfont=sf]{subfig}
\else
\usepackage[caption=false, font=footnotesize]{subfig}
\fi
\makeatletter

\newtheorem{definition}{Definition}

\makeatother

\def\BibTeX{{\rm B\kern-.05em{\sc i\kern-.025em b}\kern-.08em
    T\kern-.1667em\lower.7ex\hbox{E}\kern-.125emX}}

\begin{document}

\IEEEoverridecommandlockouts
\IEEEpubid{\begin{minipage}[t]{\textwidth}\ \\[1pt]
        \centering\footnotesize{\begin{tcolorbox}[left = 0.5mm, right = 0.5mm, top = 0.5mm, bottom = 0.5mm]\copyright This work has been submitted to the IEEE for possible publication. Copyright may be transferred without notice, after which this version may no longer be accessible.\end{tcolorbox}}
\end{minipage}}

\title{Learning to Transfer for Evolutionary Multitasking}
\author{Sheng-Hao Wu, Yuxiao Huang, Xingyu Wu*, \\Liang Feng*, \emph{Senior Member}, \emph{IEEE}, 
Zhi-Hui Zhan*, \emph{Fellow}, \emph{IEEE}, Kay Chen Tan, \emph{Fellow}, \emph{IEEE} \vspace{-0.8cm}}

\markboth{Journal of \LaTeX\ Class Files,~Vol.~18, No.~9, September~2020}%
{How to Use the IEEEtran \LaTeX \ Templates}

\maketitle

\begin{abstract}
  Evolutionary multitasking (EMT) is an emerging approach for solving multitask optimization problems (MTOPs) and has garnered considerable research interest. The implicit EMT is a significant research branch that utilizes evolution operators to enable knowledge transfer (KT) between tasks. However, current approaches in implicit EMT face challenges in adaptability, due to the use of a limited number of evolution operators and insufficient utilization of evolutionary states for performing KT. This results in suboptimal exploitation of implicit KT's potential to tackle a variety of MTOPs. To overcome these limitations, we propose a novel Learning to Transfer (L2T) framework to automatically discover efficient KT policies for the MTOPs at hand. Our framework conceptualizes the KT process as a learning agent's sequence of strategic decisions within the EMT process. We propose an action formulation for deciding when and how to transfer, a state representation with informative features of evolution states, a reward formulation concerning convergence and transfer efficiency gain, and the environment for the agent to interact with MTOPs. We employ an actor-critic network structure for the agent and learn it via proximal policy optimization. This learned agent can be integrated with various evolutionary algorithms, enhancing their ability to address a range of new MTOPs. Comprehensive empirical studies on both synthetic and real-world MTOPs, encompassing diverse inter-task relationships, function classes, and task distributions are conducted to validate the proposed L2T framework. The results show a marked improvement in the adaptability and performance of implicit EMT when solving a wide spectrum of unseen MTOPs.
\end{abstract}
\begin{IEEEkeywords}
Evolutionary multitasking, reinforcement learning, learning to optimize.
\end{IEEEkeywords}

\section{Introduction}

\IEEEPARstart{I}{n} the real world, complex optimization problems abound. Traditional numerical optimization algorithms often struggle to solve these problems efficiently due to their non-convexity and black-box properties. Against this backdrop, evolutionary computation (EC) \cite{zhan2022survey} algorithms have been widely adopted to tackle such challenging optimization problems. However, these algorithms are typically designed to solve one task at a time, which can be computationally inefficient when dealing with a multitude of intertwined tasks that potentially share some similarities and commonalities. Recognizing this limitation, a more advanced search paradigm, known as evolutionary multitasking (EMT) \cite{gupta2015multifactorial, tan2021evolutionary}, has recently emerged. Unlike traditional EC algorithms, EMT does not handle each task in isolation. Instead, it seeks to solve multiple optimization tasks simultaneously by leveraging the inherent complementarities and sharing useful information between the tasks to improve search performance. The problem of solving multiple tasks concurrently is called the multitask optimization problem (MTOP) while the technique of sharing useful information is referred to as knowledge transfer (KT) \cite{wu2023transferable}. As the core of the EMT, the KT process plays a vital role in achieving satisfactory optimization performance.

Typically, existing EMT algorithms can be classified into two categories according to the design of KT including the implicit EMT \cite{gupta2015multifactorial} and the explicit EMT \cite{feng2018evolutionary}. The implicit EMT algorithms employ the evolution operators for conducting implicit KT while explicit EMT algorithms involve an explicit learning and transformation process to perform knowledge extraction for KT. The implicit EMT is efficient and straightforward because it implements KT based on search operators with a small computational overhead. Therefore, implicit EMT has been successfully applied to solve various complex optimization problems \cite{chen2020evolutionary, feng2020explicit}.

Despite its promise, a proper design of the KT process in implicit EMT requires considering the issues of \emph{when to transfer} and \emph{how to transfer} to ensure transfer effectiveness. The ``when to transfer'' issue refers to dynamically adapting the implicit KT intensity while the ``how to transfer'' issue refers to the process of extracting knowledge by evolution operators such as crossover \cite{gupta2015multifactorial} from the source task. However, existing designs of KT intensity adaptation rely on limited information of the current evolution state such as the feedback of KT \cite{chen2019adaptive} and estimated inter-task similarity \cite{chen2019adaptive} based on manually defined heuristic rules. This insufficient use of evolution information can lead to suboptimal designs of KT. Moreover, most KT processes in implicit EMT rely on a single \cite{gupta2015multifactorial} or just a few \cite{zhou2020toward} evolution operators. The optimization tasks in practical MTOPs often display a wide array of unique characteristics, necessitating a diverse range of evolution operators for effective adaptation. This constrained use of evolution operators could hinder the ability of implicit KT to adapt to new and unseen MTOPs with varied task distributions. Therefore, the challenge lies in how to better adapt the KT intensity and the evolution operator within the KT process to suit the specific needs of MTOPs that have tasks originating from a particular distribution. Additionally, designing a suitable KT process is usually human expertise-driven and takes considerable trial-and-test efforts on the problem set of interest. This human-dependent design process hinders users from applying EMT to solve a wide range of real-world MTOPs.

Given the above challenges discussed, it is apparent that a more adaptive and automatic KT process is needed. To this end, we take cues from the \emph{learning to optimize} \cite{ma2024metabox} in the meta-learning domain to address these challenges. Learning to optimize is a kind of technique aiming to automatically discover how to perform optimization from the optimization experience gained on a problem set by machine learning methods such as reinforcement learning (RL). In specific, we propose a novel Learning to Transfer (L2T) framework in this paper to address these challenges. First, we identify that the KT process along the implicit EMT process can be formulated as a Markov decision process (MDP) to maximize the transfer quality and overall search performance. Based on this formulation, the KT process can be regarded as the sequential decisions on when and how to transfer made by an agent associated with certain parameters along the implicit EMT process. By parameterizing the agent as an actor-critic network, we proposed to learn well-performing implicit KT policy by proximal policy optimization (PPO) \cite{schulman2017proximal}. Specifically, PPO serves to train the network parameters so that the agent can maximize the transfer quality and search efficiency. Finally, the learned agent can be seamlessly incorporated with different EC algorithms to solve both the seen and the unseen MTOP instances. To our knowledge, little research has been paid to improve the adaptability and automation of the KT process by jointly addressing when to transfer and how to transfer with the RL technique.

The contributions of this paper are two-fold. First, we introduce an L2T framework that refines the adaptability of the KT process within the implicit EMT process. This framework enhances the ability of implicit EMT to effectively address a broad spectrum of MTOPs, accommodating an array of task similarities and relationships. Second, we propose a test suite with diverse task distributions and function characteristics to evaluate the adaptability of the EMT algorithms and conduct comprehensive experiments to validate the effectiveness of the proposed L2T on this test suite. 

The rest of this paper is organized as follows. Section II provides an overview of EMT, related works on KT for implicit EMT, and learning to optimize. Section III introduces the proposed L2T framework and its implementation in detail. Section IV presents the experimental studies to evaluate the performance and verify the effectiveness of the proposed L2T framework. Finally, the concluding remarks and future works are given in Section V.

\section{Preliminaries}
\subsection{Evolutionary Multitasking}
The EMT is a paradigm for optimizing multiple tasks with EC algorithms as the base solver such as differential evolution (DE) \cite{li2021meta} incorporated with a KT process. The implicit EMT differs from the explicit EMT by using evolution operators to perform KT. We briefly review the evolution process by the base solver DE and the probabilistic sampling model of KT-based offspring generation in EMT, as they form the cornerstone of our proposed method.

Let $K$ denote the number of tasks to be optimized in an MTOP instance. At generation $g$, to produce the $i$-th offspring $u_{k,i}$ by the DE algorithm for the $k$-th optimization task $f_k$ ($k=1,...,K$), two steps including mutation and crossover are carried out. Although many mutation and crossover operators have been proposed in DE, we employ the widely used ``DE/rand/1'' mutation and binomial crossover for brevity. The mutation operator first generates a trial vector $v_{k,i}$ as
\begin{equation}
  v_{k,i}=\underbrace{X_{k,r_1}}_{\text{base vector}} +F\cdot \underbrace{(X_{k,r_2}-X_{k,r_3})}_{\text{differential vector}}\text{,}
  \label{eq: de_mutation}
\end{equation}%
where $X$ is the parent population at generation $g$, $F$ is the scaling factor in DE, and $r_1, r_2, r_3$ are distinct indices randomly selected from $\{1,...,N\}$, where $N$ denotes the population size. In this mutation operation, $X_{k,r_1}$ is called the \emph{base vector} and $(X_{k,r_2}-X_{k,r_3})$ is called the \emph{differential vector}. Then an offspring solution $u_{k,i}$ is generated as
\begin{equation}
  u_{k,i,d} = 
  \begin{cases} 
  v_{k,i,d} & \text{if } rand_d(0, 1) \leq CR \\& \text{ or } d=rand_i(1,D)\text{,} \\
  X_{k,i,d} & \text{otherwise,}
  \end{cases}
  \label{eq: de_crossover}
\end{equation}%
where $d$ is the dimension index, $D$ denotes the task dimensionality, and $CR$ is a parameter to control crossover rate. If the offspring is only generated by DE operators (Eq. (\ref{eq: de_mutation})-(\ref{eq: de_crossover})), the evolution becomes the single-task optimization since no additional information from other tasks is injected. Hence, the KT operation is introduced in EMT by combining information from other tasks. According to \cite{gupta2017insights}, the KT process of generating offspring $x_k$ for target task $f_k$ with the base solver $\mathcal{A}$ (e.g., DE) can be modelled as
{
\setlength{\abovedisplayskip}{3pt}
\setlength{\belowdisplayskip}{3pt}
\begin{equation}
    \begin{split}
    p(x_k|\{\mathcal M_j\}_{j=1}^K,\mathcal{A}) = & \alpha_kp(x_k|\mathcal{M}_k,\mathcal{A}) + \\ &  \sum_{j\neq k} \alpha_jp(x_k|\mathcal{M}_j,\mathcal{A})\text{,}
    \end{split}
    \label{eq: original_sample_mdl}
\end{equation}
}%
where $\mathcal{M}$ denotes the experience collected along the search process until the current generation $g$. In the case of DE, $\mathcal{M}$ is represented as the current population $X_k$ and its fitness $Y_k$ while $\mathcal{A}$ is represented as the mutation and crossover operator. For simplicity, we assume all tasks use the same base solver $\mathcal{A}$. Moreover, $\alpha_k$ is a parameter to balance between self evolution (i.e., using the base solver with target task-only information) and KT (i.e., combining the information from source tasks) to generate offspring. Different implicit KT designs on controlling $\alpha_k$ vary from fixed \cite{gupta2015multifactorial} to adaptive \cite{bali2019multifactorial}.

A limitation of this model is that it only considers the mixture of the transferred solutions that are directly sampled from the source task distribution. However, as revealed in the previous study \cite{lin2023ensemble}, performing domain adaptation to transform the source task distribution can enhance the KT utility. Hence, we can rewrite the model to a more general form to cover the transformation-based KT process as
{
\begin{equation}
  \begin{split}
  p(x_k|\{\mathcal M_j\}_{j=1}^K,\mathcal{A},\mathcal{K} )=& \alpha_k p(x_k|\mathcal{M}_k,\mathcal{A}) + \\ &  (1-\alpha_k) p(x_k|\{\mathcal M_j\}_{j=1}^K,\mathcal{A},\mathcal{K} )\text{,}
  \end{split}
  \label{eq: improved_sample_mdl}
\end{equation}
}%
herein, $\alpha_k$ is a parameter concerning \emph{when to transfer} for a task  $f_k$. $\mathcal{K} $ typically is a function that takes the information $\{X_j\}_{j=1}^K$ of all the tasks as input and undergoes a knowledge extraction to produce a transferred offspring $v_{k,i}$, i.e., $\mathcal{K} :\{X_j\}_{j=1}^K\rightarrow v_{k,i}$. Different designs of KT operators can be regarded as an implementation of $\mathcal{K} $. Hence, $\mathcal{K} $ is a function concerning \emph{how to transfer}. Moreover, we provide a brief literature review on existing KT methods in addressing the issues of \emph{when to transfer} and \emph{how to transfer}. Due to the page limits, the discussion on existing KT designs is given in Section S.I in the supplementary material \cite{wu2024learning}.

Despite the development of various KT designs booming, the issue of \emph{limited adaptability} of the KT process still requires research attention. The adaptability of the KT process is defined as the average optimization performance on a set of unseen MTOPs that haven't been seen during the development and tuning process of the EMT algorithm. We conduct experiments to examine the adaptability of existing KT processes \cite{xu2021evolutionary, li2021meta, feng2017empirical, jin2019study, wang2021solving} that have different specially focused designs on the issues of when to transfer and how to transfer (refer to Section IV-B). We find that the current implicit EMT algorithms exhibit a restricted capacity for adapting to different MTOPs with diverse and complex task distributions, warranting further study. Moreover, a more automatic KT process that can learn good policies on deciding when and how to transfer based on the MTOPs at hand is expected.

\subsection{Learning to Optimize}

Learning to optimize is an active research branch that uses machine learning techniques to improve optimization efficiency. Among various machine learning techniques, RL is a popular one for learning to perform the search properly in each step of the optimization process. The evolutionary process can be viewed as a sequential decision-making problem taking multiple sampling and distribution update steps to approach the optima. Many researchers have seen the potential of RL in assisting the EC to tackle this decision-making problem, and proposed some RL methods to automatically design EC structure \cite{yi2022automated}, control EC parameters \cite{sun2021learning}. Additionally, several studies have employed artificial neural networks to assimilate and leverage evolutionary knowledge by predicting the population center \cite{zhan2022learning} and evolutionary direction \cite{jiang2023knowledge} throughout the search process, yielding promising outcomes. It should be noted that as a component of EMT, the goal of the KT process is to make the proper decisions on when and how to perform transfer between tasks to achieve the best optimization performance over all the tasks. Intuitively, this goal shares certain similarities with existing works on using RL to automate evolutionary process \cite{zhang2021learning}. Therefore, inspired by learning to optimize, we proposed the L2T framework to automatically learn well-performing KT policy for implicit EMT.

\section{Proposed Method}
\subsection{L2T Workflow}

\begin{figure}[htbp]
  \centering
  \includegraphics[width=1\columnwidth]{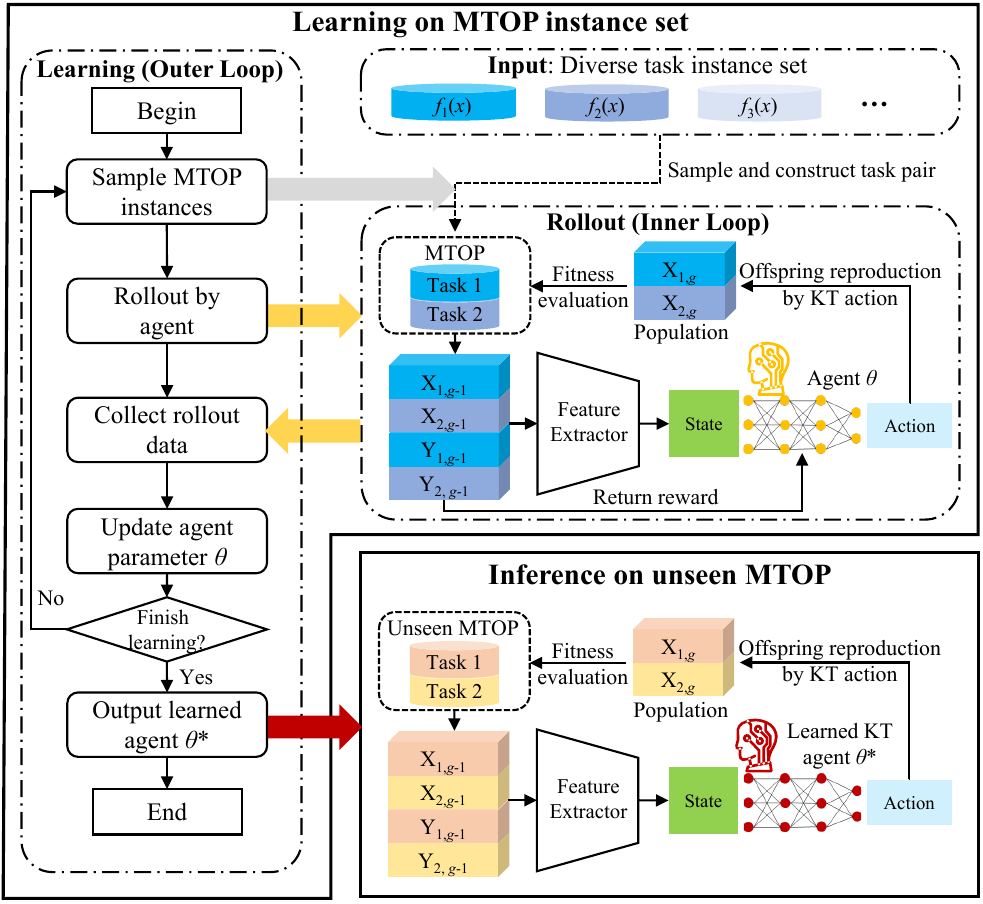}
  \caption{Diagram of the proposed L2T framework.}
  \label{fig: algorithm}
  \vspace{-0.4cm}
\end{figure}

The diagram of the proposed L2T framework is shown in Fig. \ref{fig: algorithm}. Generally, L2T contains two stages, i.e., the learning and the utilization stage. The input of the learning stage is a task instance set that contains diverse optimization tasks that can be used to sample and construct MTOP instances. The learning stage contains an outer loop for learning the KT policy and an inner loop for interaction between the agent and environment called rollout. In the learning stage, the EMT environment continuously generates new MTOP instances for an agent to solve. In the rollout process, the agent undergoes an EMT process by taking actions to produce offspring. Specifically, for each time step ${g}$ (i.e., a generation in the EC process), the agent receives a state after performing feature extraction on the populations $X_{g-1}$ and fitness $Y_{g-1}$. Then the agent outputs an action to perform KT-based sampling to obtain offspring population $X_{g}$. A complete rollout process is called an episode. In an episode, the state, action, and reward of each time step will be collected as rollout data. After a certain number of episodes has been taken, the agent updates the parameter $\theta$ according to the RL algorithm. When the learning is finished, the agent with optimized parameters $\theta^*$ can be directly plugged and used in EMT algorithms to solve new and unseen MTOPs. The major components of the L2T framework include the definition of the RL components (Section III-B), how to learn the agent and how the agent is plugged into the EMT algorithm in the utilization stage (Section III-C). It should be noted the main computational cost is consumed in the learning stage and the time complexity of L2T in the utilization stage is small, which is analyzed in Section III-D.

\subsection{L2T Component Design}
\subsubsection{MDP Formulation of KT Process}
We first formulate the KT process within the implicit EMT as an MDP problem that takes high-valued actions (i.e., deciding when and how to transfer) to maximize the cumulative reward (i.e., yielding the best results when the optimization finishes). Normally, an MDP can be represented as a tuple $(a, s, r, \pi, p_{tr})$ where $a$ denotes the action, $s$ denotes the state, $r$ denotes the reward function, $\pi$ denotes the probability distribution of the agent action parameterized by $\theta$, and $p_{tr}$ denotes the state transition probability. The objective is to find an optimal agent $\pi^*(a|s;\theta)$ parameterized by $\theta$ to realize $\max_{\pi}\mathbb{E}_{\tau} [p(\tau;\pi)R(\tau)]$, where $\tau=(s_0,a_0,r_0,...,s_{T},a_{T},r_T)$ denotes the realization of a trajectory, $T$ denotes the time horizon (i.e., maximum generations in EMT), and $R(\tau)=\sum_{t=0}^Tr_{t}$ denotes the cumulative reward. The probability of a trajectory $\tau$ is affected by $\pi$ and $p_{tr}$, i.e., $p(\tau;\pi)=\prod_{t=0}^T \pi(a_t|s_t;\theta)p_{tr}(s_{t+1}|s_t,a_t)$. The formulations of action $a$, state $s$, reward $r$, agent structure $\pi(a|s;\theta)$, and the environment dynamics $p_{tr}$ will be introduced in the following.

\begin{table}[htbp]
  \setlength{\abovecaptionskip}{0cm}
  \setlength{\belowcaptionskip}{-0.2cm}
  \centering
  \setlength{\tabcolsep}{2pt} 
  \caption{Interpretation of Different KT Actions}
      \begin{tabular}{m{2em}<{\centering}m{2em}<{\centering}m{2em}<{\centering}m{8em}<{\centering}m{15em}<{\centering}}
      \toprule
      \multicolumn{3}{c}{Action parameters} & \multicolumn{2}{c}{Translated operation} \\
      {\textit{$a_{k,1}$}} & {\textit{$a_{k,2}$}} & {\textit{$a_{k,3}$}} & When to transfer & How to transfer \\
      \midrule[0.06em] 
      0    & / & / & Without KT & Self evolution by $\mathcal{A}$ (e.g., Eq. (\ref{eq: de_mutation})) \\
      \midrule[0.06em]
      \multirow{4}[14]{*}{1}     & 0     & 1     & KT probability=0.5  & Transfer differential vector (Eq. (\ref{eq: differential vector transfer}))   \\
      \cmidrule[0.06em]{2-5}          & 0.5   & 0     & KT probability=0.5  & Crossover on source and target solutions (Eq. (\ref{eq: arithmetic_crossover})) \\
      \cmidrule[0.06em]{2-5}          & 1     & 0     & KT probability=0.5  & Transfer source solutions as base vector (Eq. (\ref{eq: base vector transfer})) \\
      \cmidrule[0.06em]{2-5}          & 1     & 1     & KT probability=0.5  & Transfer sampled solution from source distribution (Eq. (\ref{eq: direct transfer}))\\
      \bottomrule
      \end{tabular}%
  \label{table: Interpretation}
\end{table}%

\subsubsection{Action Formulation}
We first briefly revisit some KT operators that are related to our design of the action space for the KT process. Some representative KT operators include linear combination-based crossover \cite{zhou2020toward}, base vector transfer \cite{jin2019study}, and differential vector transfer \cite{feng2017empirical}. The linear combination-based crossover including arithmetic crossover \cite{zhou2020toward} and simulated binary crossover (SBX) \cite{gupta2015multifactorial} is to perform interpolation on the target and source solutions to generate offspring $v_{k,i}$ as
\begin{equation}
  v_{k,i}=\lambda \cdot X_{k,r_1}+(1-\lambda)\cdot X_{j,r_2}\text{,}
  \label{eq: arithmetic_crossover}
\end{equation}%
where $j\neq k$ is a randomly selected source task index from $\{1,...,K\}$ and $\lambda\in[0,1]$ is a random variable that is drawn from a predefined distribution. The base vector transfer is
\begin{equation}
  v_{k,i}=X_{j,r_1}+F\cdot (X_{k,r_2}-X_{k,r_3})\text{,}
  \label{eq: base vector transfer}
\end{equation}
while the differential vector transfer is 
\begin{equation}
  v_{k,i}=X_{k,r_1}+F\cdot (X_{j,r_2}-X_{j,r_3})\text{.}
  \label{eq: differential vector transfer}
\end{equation}
Moreover, the direct transfer \cite{wang2021solving} that uses the sampled solution from the source population is a well-known KT operator as
\begin{equation}
  v_{k,i}=X_{j,r_1}+F\cdot (X_{j,r_2}-X_{j,r_3})\text{.}
  \label{eq: direct transfer}
\end{equation}
It is widely studied that different KT operators have different transfer biases and favor different kinds of task similarity and evolution states \cite{lin2023ensemble}. This inspires us to make use of the complementarity of different KT operators and devise a more generalized KT action representation that can adapt to diverse evolution states and task similarities. Hence, combining the KT probabilistic model in Eq. (\ref{eq: improved_sample_mdl}) and existing KT operators, we derive the sampling model for an offspring $v_{k,i}$ at generation $g$ as
\begin{equation}
  \begin{split}
  v_{k,i}\sim & (1-0.5 \cdot a_{k,1}) \cdot p(v_{k}|X_k,\mathcal{A})+ \\ & 0.5 \cdot  a_{k,1} \cdot p(v_{k}|\{X_j\}_{j=1}^K,\mathcal{A},\mathcal{K} )\text{,}
  \end{split}
  \label{eq: derived_sample_mdl}
\end{equation}
and $v_{k} \sim p(v_{k}|\mathcal{K}  (\{X_j\}_{j=1}^K)) $ is defined as
\begin{equation}
  \begin{split}
    v_{k}=& (1-a_{k,2})\cdot X_{k,r_1} + a_{k,2}\cdot X_{j,r_2} +\\ & F\cdot (1-a_{k,3}) \cdot (X_{k,r_3} - X_{k,r_4}) + \\ & F\cdot a_{k,3} \cdot (X_{j,r_5} - X_{j,r_6})\text{,}
  \end{split}
  \label{eq: sample_mutation_vector}
\end{equation}
where $r_1,r_2,r_3,r_4,r_5,r_6$ are randomly drawn from $\{1,...,N\}$. $a_{k,1}, a_{k,2}, a_{k,3}\in[0,1]$ are three continuous action parameters output by an agent for task $f_k$ and can be varied among generations. After generating a trial vector $v_{k,i}$, the binomial crossover operation is put forward to produce the final offspring $u_{k,i}$. According to the analysis in \cite{bali2019multifactorial}, we impose the restriction on the probability of self evolution by base solver $\mathcal{A}$ to be larger than 0.5 under truncated selection mechanism, i.e., $\alpha_k=1-0.5\cdot a_{k,1}\geq 0.5$, to guarantee convergence. The main feature of this action formulation is that by setting different $a_{k,i}, a_{k,2}, a_{k,3}$, the mutation operator in Eq. (\ref{eq: de_mutation}) and existing KT operators Eq. (\ref{eq: arithmetic_crossover})-(\ref{eq: direct transfer}) can be recovered from Eq. (\ref{eq: sample_mutation_vector}). That is, the above-mentioned self evolution operator and KT operators are a subset of the proposed action space. The detailed interpretation of different KT actions is shown in Table \ref{table: Interpretation}. Note that since actions $a_{k,1}, a_{k,2}, a_{k,3}$ are continuous, setting action values within $(0,1)$ refers to the combination of multiple KT operators. Intuitively, we expect that more candidate actions of linearly combining KT operators can provide more diverse behaviors and potentially improve the capacity of the KT process to adapt to more diverse MTOPs.

\begin{table*}[htbp]
  \vspace{-0.4cm} %
  \setlength{\abovecaptionskip}{0cm}
  \setlength{\belowcaptionskip}{-0.2cm}
  \centering
  \setlength{\tabcolsep}{2pt} 
  \caption{State Features for the Input of the Learnable agent}
    \begin{tabular}{cccc}
    \toprule
    Feature type & Notation and definition & Range & Meaning \\
    \midrule
    \multirow{4}[2]{*}{Common feature} & $O_{c,1}=g/G_{\max}$ & [0, 1] & Ratio of current generation $g$ \\
          & $O_{c,2}=d(x^{*}_{1},x^{*}_{2})/\sqrt{D} $ & [0, 1] & Distance between best-found solutions till current generation $g$ \\
          & $O_{c,3}=d(\mu_{1},\mu_{2})/\sqrt{D} $ & [0, 1] & Distance between first-order statistics of current population \\
          & $O_{c,4}=d(\sigma_{1},\sigma_{2})/\sqrt{0.5D} $ & [0, 1] & Distance between second-order statistics of current population \\
    \midrule
    \multirow{7}[2]{*}{Task-specific feature} & $O_{t,k,1}=n_{k,stag}/G_{\max}$ & [0, 1] & Ratio of the number of stagnating generations \\
          & $O_{t,k,2}=flag_{k,\rm{improved}}$& [0, 1] & Whether best solution is improved in last generation \\
          & $O_{t,k,3}=q_{k,KT}$ (Eq. (\ref{eq: transfer quality}))   & [0, 1] & Transfer quality of the last knowledge transfer action \\
          & $O_{t,k,4}=mean(std(X_{k}))$ & [0, 1] & Average deviation of the current population over all dimensions \\
          & $O_{t,k,5}=a_{k,1,g-1}$    & [0, 1] & Last KT action taken for deciding when to transfer \\
          & $O_{t,k,6}=a_{k,2,g-1}$     & [0, 1] & Last KT action taken for deciding how to transfer \\
          & $O_{t,k,7}=a_{k,3,g-1}$     & [0, 1] & Last KT action taken for deciding how to transfer \\
    \bottomrule
    \end{tabular}%
  \vspace{-0.4cm}
  \label{table: State Features}
\end{table*}%

\subsubsection{State Representation}
Now we proceed to define what information should be fed into the agent for supporting it to make high-quality decisions. A naive way is to directly feed all the populations and fitness of tasks at generation $g$ to the agent without any processing. However, this input yields a large dimension, i.e., $K \cdot N\cdot (D+1)$ where $K$ denotes the number of tasks, $D+1$ contains the solution dimension and fitness dimension. Since not all information in the population is useful, the large input may distract the agent from attending to useful information for decision-making. Therefore, we extract features from the populations to make the agent easier to learn. Existing adaptive KT designs prefer using feedback of KT \cite{xu2021evolutionary} and estimated inter-task similarities \cite{chen2019adaptive} as features for adapting intensity parameters. By combining these two ideas and adding more informative features, the proposed state features are given in Table \ref{table: State Features}. The incorporated features can be divided into common features $O_c$ and task-specific features $O_t$. The common features are shared among all the tasks while the task-specific features are calculated for each task independently. The common features include the timestamps of the current generation $O_{c,1}=g/G_{\max}$, where $g$ is the current generation and $G_{\max}$ is the maximum generations and three distance metrics $O_{c,2}, O_{c,3}, O_{c,4}$ to estimate task similarity where $d(\cdot,\cdot)$ is the Euclidean distance. The task-specific features include the stagnation state $O_{t,1}=n_{stag}/G_{\max}$ where $n_{k,stag}$ is the number of stagnating generations, the improvement state $O_{t,2}=flag_{k,\rm{improved}}\in \{0,1\}$  indicating whether the best solution has been improved in last generation, the transfer quality of last KT action $O_{t,3}=q_{k,KT}$, the average deviation of the current population $O_{t,4}=mean(std(X_{k}))$, where $mean(\cdot)$ and $std(\cdot)$ respectively are the mean and standard deviation operations, and the last KT action $O_{t,5},O_{t,6},O_{t,7}$. Specifically, given the fitness of the population at $g$-th generation $Y_{k,g}=\{y_{k,1},...,y_{k,N}\}$ and the offspring produced by KT, denoted as $Y_{k,\mathcal{K}}=\{y_{k,\mathcal{K},1},...,y_{k,\mathcal{K},N_{k,KT}}\}$, where $N_{k,KT}$ denotes the number of KT-generated solutions for the task $f_k$, the transfer quality is calculated as
\begin{equation}
  q_{k,KT} = 
  \begin{cases} 
  \frac{\sum_{i = 1}^{N_{k,KT}} \left\lvert \{y_{k,g}\in Y_{k,g}:y_{k,\mathcal{K},i}<y_{k,g}\}  \right\rvert}{N_{k,KT} \cdot N} & \text{if } N_{k,KT}>0 \text{,} \\
  0 & \text{otherwise,}
  \end{cases}
  \label{eq: transfer quality}
\end{equation}%
where $|\cdot|$ is the cardinality of a set.
To make the magnitude of different features compatible, we normalize them to the range $[0,1]$. Hence, given $K$ tasks, the state dimension after the feature extraction is $\left\lvert O_c\right\rvert + \left\lvert O_t\right\rvert=4+7 \cdot K$.

\subsubsection{Reward Design}
The reward serves as a temporary signal for incentivizing the agent to learn good policies. The objective for the agent in EMT is to achieve the targeted optimization accuracy $\xi$ when the generation $g$ has been reached, i.e., $\mathbb{I} (f_k(x_{k,g}^*)-f_k^*<\xi)$, where $x_{k,g}^*$ is the best-found solution by the EMT algorithm until generation $g$, $\mathbb{I}(\cdot)$ is the indicator function, and $f_k^*=\min_x f_k(x)$ is the optimal objective value. Simply giving the reward to the agent at the end of a rollout process if the target accuracy is reached will lead to the sparse reward issue \cite{khadka2018evolution}, which is difficult for the agent to learn. To alleviate this issue, we aggregate two additional reward terms, namely convergence gain $r_{k,conv}$ and knowledge transfer gain $r_{k,KT}$. The reward over all tasks is defined as
\begin{equation}
  r=\sum\nolimits_{k=1}^{K}r_k,
\end{equation}%
where $r_k$ defines the sub-reward on task $f_k$ as
\begin{equation}
    r_k=\beta_1 \cdot r_{k,conv} + \beta_2 \cdot r_{k,KT} + \beta_3 \cdot \mathbb{I} (f_k(x_{k,g}^*)-f_k^*<\xi) \text{,} 
    \label{eq: reward function}
\end{equation}%
where 
\begin{equation}
    r_{k,conv}=-(f_k(x_{k,g}^*)-f_k^*)/(f_k(x_{k,1}^*)-f_k^*) \text{,}
    \label{eq: reward convergence} 
\end{equation}%
and
\begin{equation}
    \begin{split}
    r_{k,KT}= & \mathbb{E} _{x_k\sim p(x|X_k,\mathcal{A} )}[f_k(x_k)] - \\ & \mathbb{E} _{x_k\sim p(x|\{X_j\}_{j=1},\mathcal{K} )}[f_k(x_k)] \text{.} 
    \end{split}
    \label{eq: reward kt} 
\end{equation}%
One important property of $r_{k,KT}$ is that it rewards the agent with a positive signal for producing a higher expectation of fitness quality by KT than the base solver $\mathcal{A}$ and a negative signal vice versa. This reward design is expected to drive the agent to balance between the risks of negative transfer and the performance gain brought by KT. In practice, $r_{k,KT}$ is approximated and implemented as the difference in the average normalized scores of KT-generated offspring solutions and that of base solver-generated offspring solutions, i.e., 
\begin{equation}
  r_{k,KT}\approx \sum\nolimits_{j=1}^{N_{\mathcal{A}}} s(y_{\mathcal{A},j})- \sum\nolimits_{j=1}^{N_{\mathcal{K}}} s(y_{\mathcal{K},j}).
\end{equation} 
$y_{\mathcal{A},j}$ denotes the fitness of $j$-th offspring produced by $\mathcal{A}$ and $y_{\mathcal{K},j}$ denotes the fitness of $j$-th offspring produced by $\mathcal{K}$. $s(y)$ denotes the score of an offspring fitness at generation $g$, and is calculated as a normalized value within $[0,1]$ over the parent population of the corresponding task, i.e., 
\begin{equation}
  s(y)=\left\lvert \{y_{g}\in Y_{g}:y<y_{g}\}  \right\rvert / N.
\end{equation}%
Notably, despite the reward requires ground-truth information about the MTOP instances such as optimal objective value, the reward is only used in the learning stage. In the utilization stage, the reward signal is not required for the learned agent.

\subsubsection{Agent and Environment}
We use the multi-layer perceptron, a feed-forward neural network denoted as $\phi(s;\theta)$ to parameterize the agent $\pi(a|s;\theta)$, given the powerful expressiveness and function fitting capability of neural networks. That is, $\theta$ is the parameters of a neural network including the weights and biases of different layers. The input of the network is the state vector by concatenating the common features and task-specific features of each task with dimensionality equal to $4+7 \cdot K$. The output of the network is the action vector containing each task's KT action with 3 dimensions, leading to a total dimension of $3 \cdot K$. Moreover, to handle the continuous action space, we employ a Gaussian sampling to obtain the action as
\begin{equation}
  \pi(a|s,\theta)=\frac{1}{\sqrt{2\pi \sigma}}\exp (- \frac{(a-\phi(s;\theta))^2}{\sigma^2})\text{,} 
  \label{eq: sample action} 
\end{equation}%
The EMT environment refers to the MTOP instances with which the agent is interacting. Specifically, given an agent $\pi(a|s;\theta)$, the state transition probability $p_{tr}(s_{t+1}|s_t,a_t)$ is mainly affected by the distribution of MTOP instances since the behaviors of EMT algorithms are driven by the returned fitness of the handling tasks. To implement the environment, we presume the existence of a task instance set, denoted as $\Theta $, where each instance is an optimization task with a known optimal objective value and is assumed to be independently and identically (i.i.d.) drawn from a task distribution $p(f)$, where $f:x\rightarrow y$ denotes a task to be optimized. We also assume that the distribution of MTOP instances solely relies on the task distribution $p(\mathcal{T})=p(f_1,...,f_K)=\prod _{k=1}^Kp(f_k)$. Therefore, we can construct an MTOP instance by randomly sampling $K$ tasks from the task instance set $\Theta $. For more details on the MTOP instance construction, refer to Section S.III in the supplementary material \cite{wu2024learning}.

\subsection{Learning and Utilizing the Learnable Agent}
\subsubsection{Learning Stage}
Having the definition of the agent and the environment, we should now determine how to efficiently learn the agent. To this end, we adopt the PPO which is good at handling continuous action space for learning the policy. In specific, the PPO algorithm contains an actor network $\pi(a|s;\theta)$ parameterized by $\theta$ to take actions and a critic network $\psi(s;\omega)$ parameterized by $\omega$ to estimate the value of the state. In our implementation, the critic network shares the same architecture as the actor network but with different parameters. The objective function of the actor network is
\begin{equation}
  L_{\pi}(\theta)=\mathbb{E}_t [\min(\rho_t(\theta)\hat{A}_t, \text{clip}(\rho_t(\theta),1-\epsilon,1+\epsilon)\hat{A}_t)] \text{,}
  \label{eq: actor objective function} 
\end{equation}%
where $\hat{A}_t^{(\gamma,\lambda)}=\sum_{l=0}^{T-t-1}(\gamma\lambda)^{l}(r_{t+l}+\psi(s_{t+l+1};\omega')-\psi(s_{t+l};\omega'))$ is the generalized advantage estimation (GAE) function, $\gamma$ is the discounted factor, $\lambda$ is a GAE parameter, and $\rho_t(\theta)=\pi(a_t|s_t;\theta)/\pi(a_t|s_t;\theta')$ is the probability ratio. 
Let $\theta',\omega'$ be the parameters of actor and critic networks that are used in collecting the latest rollout data buffer. The objective function of the critic network is
\begin{equation}
  L_{\psi}(\omega)=-\sum\nolimits_{t=1}^T(\hat{A}_t^{(\gamma,\lambda)}+\psi(s_t;\omega')-\psi(s_t;\omega))^2 \text{.}
  \label{eq: critic objective function} 
\end{equation}%
Then we can do gradient ascent to update the actor-critic network. The rollout process by the agent with base solver DE is given in \textbf{Algorithm 1}. A rollout process is terminated until the maximum rollout generations, denoted as $G_{\rm{roll}}$ is reached. In our implementation, we only consider MTOP instances with two tasks ($K=2$). That is, the input MTOP instance is a task pair. Notably, our framework can be easily extended to solve MTOPs with more than two tasks by randomly pairing the tasks into multiple sub-MTOPs and using the learned agent along with a DE solver to solve them. The main functionality of the rollout is to collect experiences for learning by interacting with the environment. To reduce environment variance for boosting learning efficiency, our implementation employs two tricks, i.e., pseudo-random initialization and deterministic KT intensity. For initialization, we bypass the usual uniform sampling at each rollout's start, opting instead to seed each task with a randomly chosen population from a pre-generated set (line 1). This primes the learning environment with a consistent baseline. With regards to KT intensity, we move away from probabilistic decisions for each offspring. Instead, we calculate the precise quota of KT solutions using $N_{k,KT}=\left\lceil 0.5\cdot a_{k,1}\right\rceil$. This allows us to construct the offspring population in two clear steps: first, by generating solutions via the DE solver, and second, by substituting $N_{k,KT}$ of these with KT-generated counterparts. This method ensures a balanced integration of KT without reliance on chance, streamlining the learning process. The learning process of the agent with PPO is given in \textbf{Algorithm 2}. Specifically, to improve the sample efficiency, we adopt multiple parallelized environments for agents to simultaneously collect rollout data. The initial population set is constructed by conducting Latin hypercube sampling independently for $N_P$ times, where $N_P$ denotes the initial population set size.

Moreover, in order to demonstrate the generality and compatibility of the proposed L2T framework under different base solvers, we choose the Genetic Algorithm (GA) as an additional example and incorporate the L2T with GA. GA involves a solution pairing process by the selection operator followed by the pairwise solution crossover, whose workflow is quite different from DE. Therefore, we implement the rollout process based on GA in a style similar to MFEA. The detailed rollout process with GA as the base solver is shown in \textbf{Algorithm S.1} in Section S.II of the supplementary material \cite{wu2024learning}.

\begin{algorithm}\small
  \caption{Agent rollout with base solver $DE$}\label{algorithm:Rollout}
  \KwIn{Task pair $ \mathcal{T}$=\{$f_1$, $f_2$\}, base solver $\mathcal{A}=DE$, maximum rollout generations $G_{\rm{roll}}$, parameterized agent $\pi(s|\theta)$, initial population set $\mathcal{P}=\{P_1,...,P_{N_P}\}$, population size per task $N$}
  \KwOut{Rollout data buffer $\mathcal{D} $}
  $\mathcal{D} = \varnothing$; \tcp{Empty rollout data buffer}
  Initialize population $X$ of size $N$ for each task by randomly selecting initial population from $\mathcal{P}$\;
  Evaluate fitness of $X$ to obtain $Y$ for each task\;
  Calculate initial state $s$ by concatenating $O_c$ and $O_t$\;
  $g=0$\;
  \While{$g<G_{\rm{roll}}$}{
    $a=\pi(s|\theta)$; \tcp{predict action by actor network}
    \ForEach{\rm{task} $f_k$}{
      Retrieve KT action parameters $a_{k,1},a_{k,2},a_{k,3}$ for task $f_k$ from $a$\;
      {Sample offspring population $U$ by Eq. (\ref{eq: de_mutation})-(\ref{eq: de_crossover})\;}
      $N_{k,KT}=\left\lceil 0.5\cdot a_{k,1}\right\rceil $; \tcp{Calculate KT quota}
      Randomly select $N_{k,KT}$ indices from $\{1,...,N\}$ to construct a index set $\mathcal{I}_{KT}=\{j_1,...,j_{N_{k,KT}}\}$\;
      \ForEach{\rm{index} $j$ in $\mathcal{I}_{KT}$}{
        Sample $v_{k,j}$ by Eq. (\ref{eq: derived_sample_mdl})-(\ref{eq: sample_mutation_vector}) with KT action parameters $a_{k,2}, a_{k,3}$ \;
        Perform crossover to obtain $u_{k,j}$ by Eq. (\ref{eq: de_crossover})\;
        Replace $j$-th solution in $U$ with $u_{k,j}$\;
      }
      Evaluate fitness $Y_k$ of $U$ on $f_k$\;
      Calculate task-specific features $O_{t,k}$ of task $f_k$\;
      Calculate reward $r_k$ based on Eq. (\ref{eq: reward function})\;
      Update populations $X_k$ by selection\;
    }
    $r=r_1+r_2$;\tcp{Sum up rewards of the tasks}
    Calculate common features $O_c$\;
    Update state $s$ by concatenating $O_c$ and $O_t$\;
    $\mathcal{D} = \mathcal{D} \cup (s,a,r) $\;
    $g=g+1$\;
  }
\end{algorithm}

\begin{algorithm}\small
  \caption{Agent learning by PPO}\label{algorithm}
  \KwIn{Task instance set $\Theta=\{f_1,...,f_K\}$, base solver $\mathcal{A}$, maximum rollout generations $G_{\rm{roll}}$, maximum time steps $T$, number of parallel environments $N_{env}$, initial population set size $N_{P}$, rollout data buffer size for PPO update $N_{buff}$}
  \KwOut{Learned agent $\pi(s|\theta^*)$}
  Pre-generate a set of populations $\mathcal{P}=\{P_1,...,P_{N_P}\}$ by Latin hypercube sampling\;
  Initialize the actor network with parameter $\theta$ and critic network with parameter $\omega$\;
  $t=0$\;
  \While{$t<T$}{
    $\mathcal{D}=\varnothing$\;
    \While{$\left\lvert\mathcal{D} \right\rvert <N_{buff}$}{
      Sample from $\Theta$ to get MTOP instances $\{\mathcal{T}_1,...,\mathcal{T}_{N_{env}}\}$\;
      Send MTOP instances to remote workers to perform rollout by the agent with parameter $\theta$; \tcp{Algorithm 1}
      Retrieve data buffer and merge $\mathcal{D}_m = \bigcup _{i}^{N_{env}}\mathcal{D}_k$\;
      $\mathcal{D} = \mathcal{D} \cup \mathcal{D}_m$\;
      $t=t+N_{env}\cdot G_{\rm{roll}}$\;
    }
    Update actor network parameter $\theta=\theta+\eta\nabla_\theta L_{\pi}$\;
    Update critic network parameter $\omega=\omega+\eta\nabla_\omega L_{\psi}$\;
    Record the best-found agent parameter $\theta^*$;
  }
\end{algorithm}

\subsubsection{Utilization Stage}
Upon completion of the agent's learning, it becomes ready for seamless integration with the DE base solver to tackle new MTOPs. The EMT algorithm we have developed, designated as Multitask Differential Evolution based on L2T (MTDE-L2T), is detailed in \textbf{Algorithm S.2}. MTDE-L2T mirrors the structure of the rollout process in \textbf{Algorithm 1}, with the notable exception of delivering optimized solutions for individual tasks. It also omits the optimization data collection and reward computation steps found in the rollout process. Importantly, MTDE-L2T accommodates varying computational resources by allowing a flexible maximum number of generations $G_{\max}$ as input during testing, which can differ from the learning stage's $G_{\rm{roll}}$. In cases where the GA serves as the base solver, we detail the Multitask GA based on L2T (MTGA-L2T) in \textbf{Algorithm S.3} in Section S.II of the supplementary material \cite{wu2024learning}.

\subsection{Time Complexity Analysis}
We analyze the time complexity of the proposed MTDE-L2T and MTGA-L2T when solving new MTOPs in the utilization stage. Since the base solver is a customizable setting for users, we leave it out and mainly discuss the worst-case complexity of the L2T-related operations, including state feature calculation, action prediction, and offspring generation by KT. We treat the operation performed on a single value such as a dimension of or a fitness of the solution as the basic operation to derive the complexity. We denote the number of tasks, dimensionality, and population size as $K,D,N$, respectively. By summing up the common features and task-specific features, the feature computation cost is $\mathcal{O}(1+D+(2N+1)D+(2N+1)D+K\cdot(1+1+N^2+(ND+D)+1+1+1))=\mathcal{O}(KN^2+KND)$. The cost of action prediction is a function of the neural network parameters, denoted as $W(4+7K,3K)$, which correlates with the input feature dimension of $4+7K$ and the output action dimension of $3K$. Normally, the number of network parameters can be considered as a linear function relating to input and output size dominated by $K$. Hence, we have the prediction cost $\mathcal{O}(K)$. The cost of offspring generation by KT is $\mathcal{O}(KND)$. Finally, by putting them all together and simplifying the non-dominant terms, we obtain the complexity of L2T-related operations as $\mathcal{O}(KN^2+KND)$.

\section{Experimental Study}
\subsection{Experimental Setup}
\subsubsection{Problem Setting and Parameter Configuration}
\begin{figure*}[!t]
  \vspace{-0.4cm}
  \centering
  \captionsetup[subfloat]{farskip=12pt,captionskip=0pt}
  \subfloat[Very small range (VS)]{\includegraphics[width=1.4in]{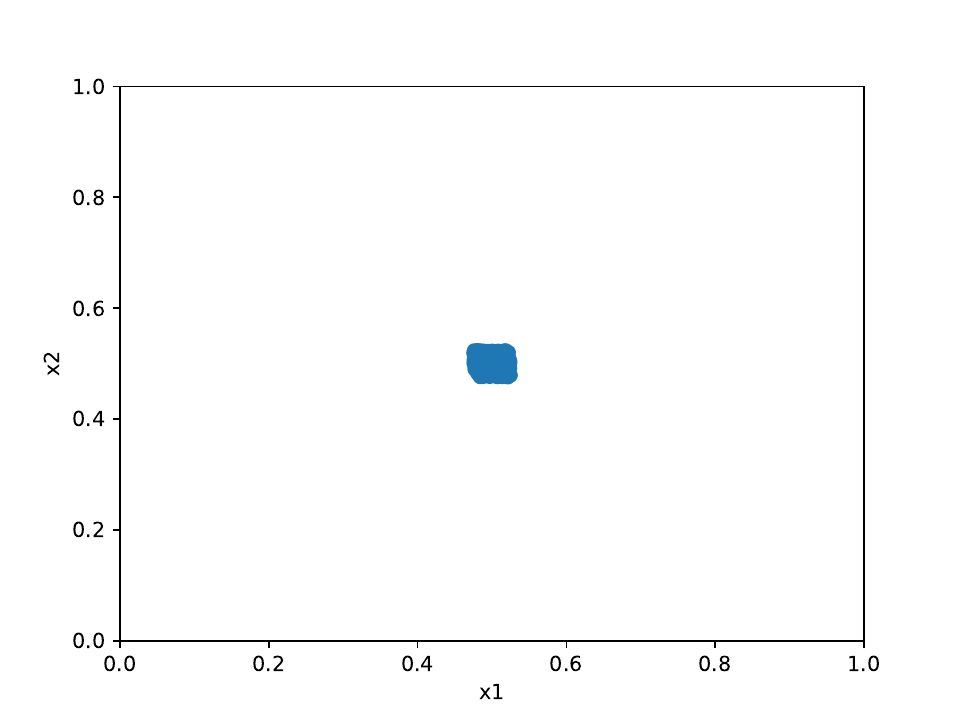}}
  \subfloat[Small range (S)]{\includegraphics[width=1.4in]{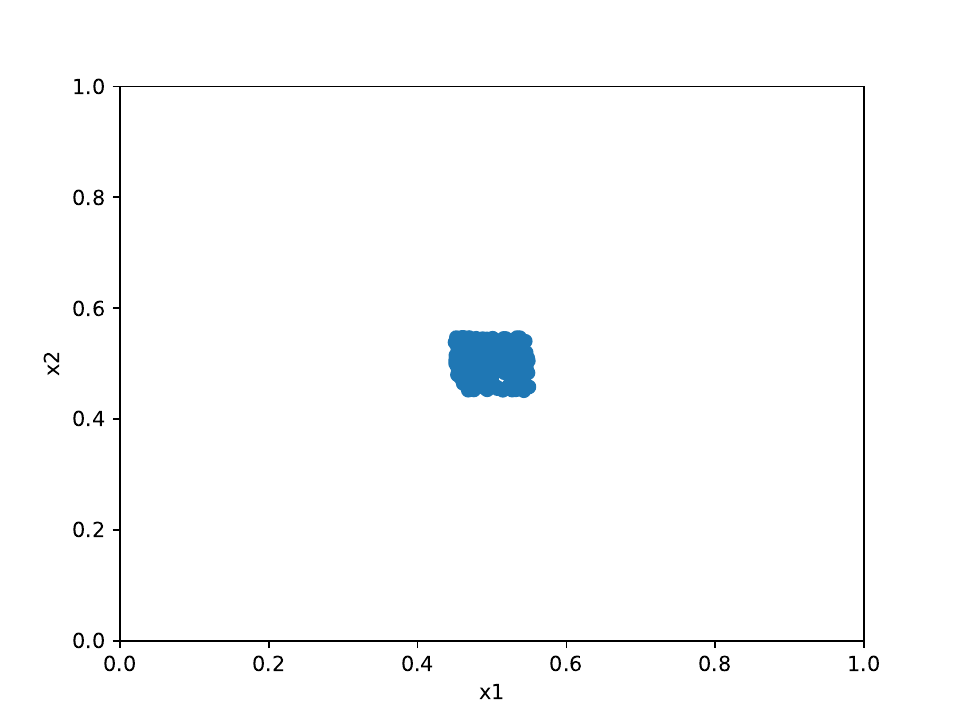}}
  \subfloat[Medium range (M)]{\includegraphics[width=1.4in]{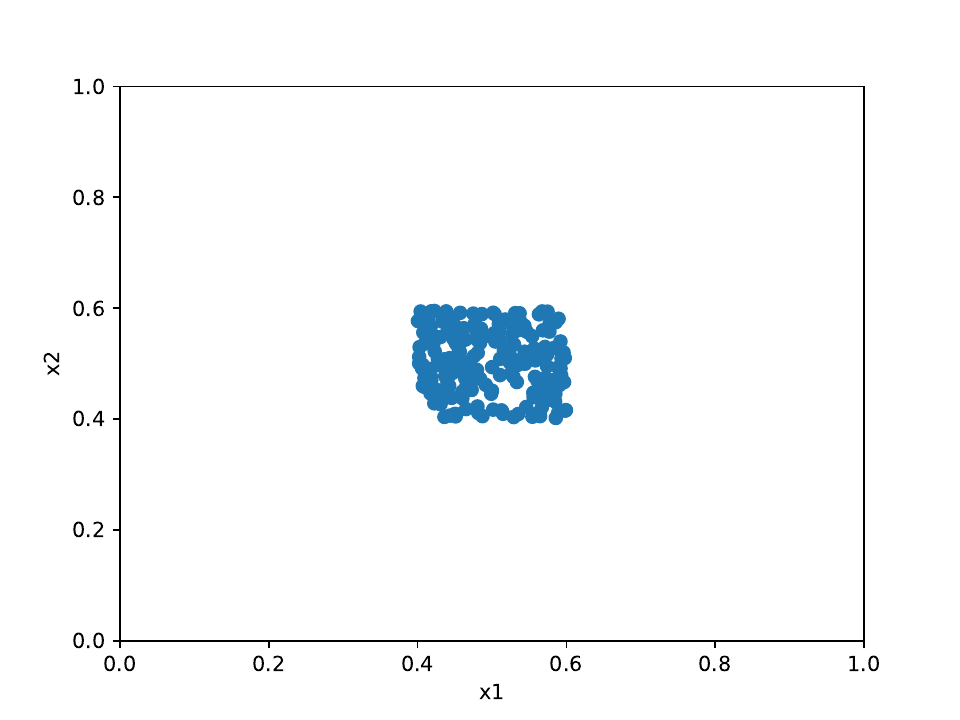}}%
  \subfloat[Large range (L)]{\includegraphics[width=1.4in]{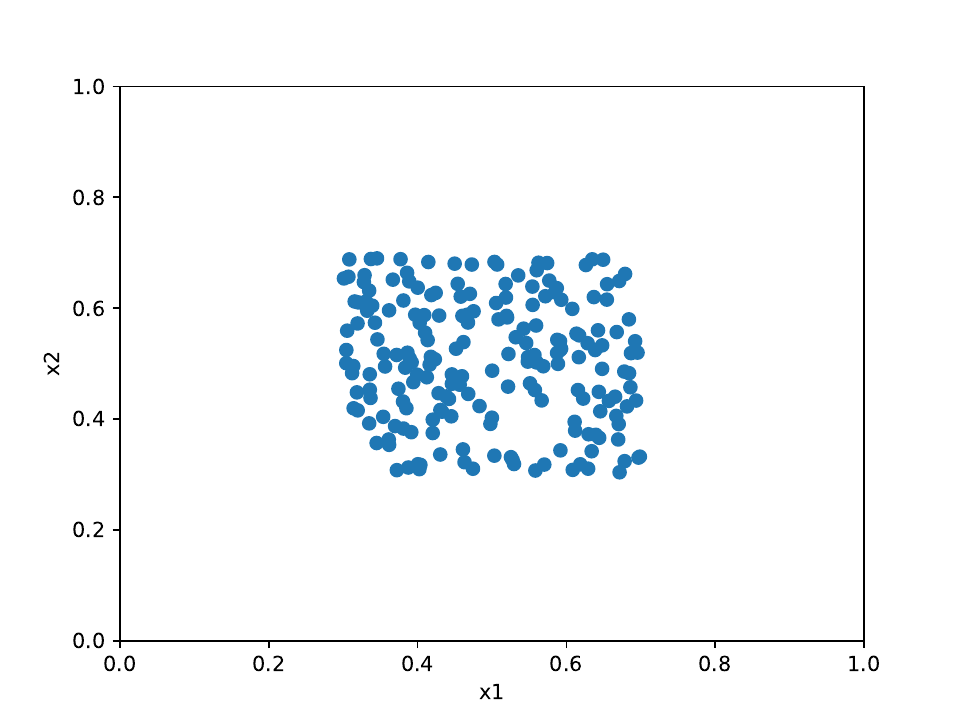}}
  \subfloat[Very large range (VL)]{\includegraphics[width=1.4in]{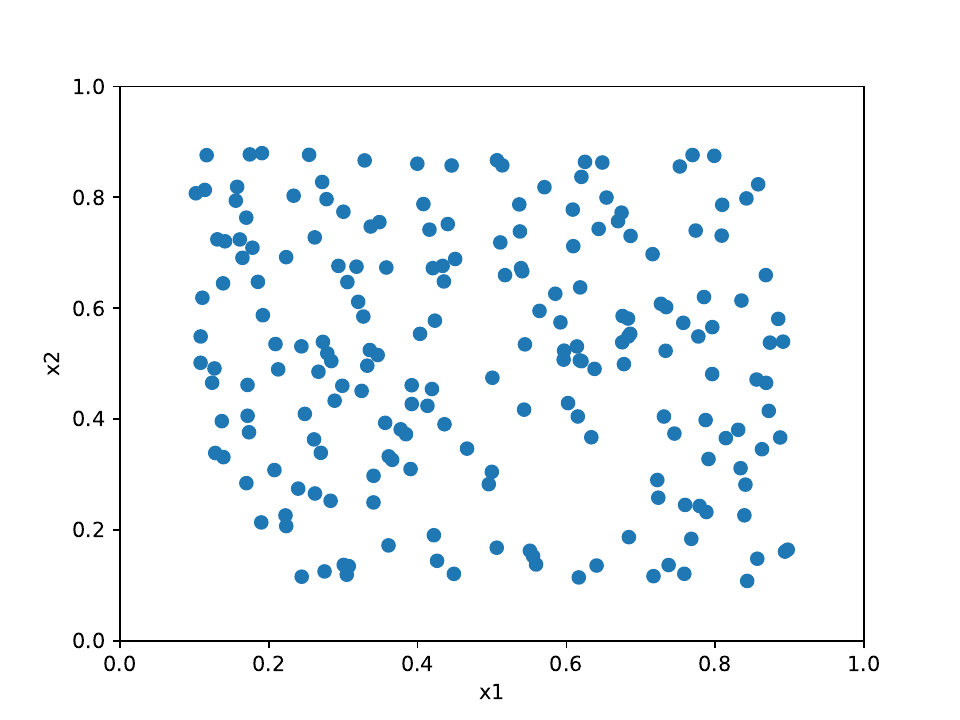}}\vspace{-4mm}
  \subfloat[Single cluster (C1)]{\includegraphics[width=1.4in]{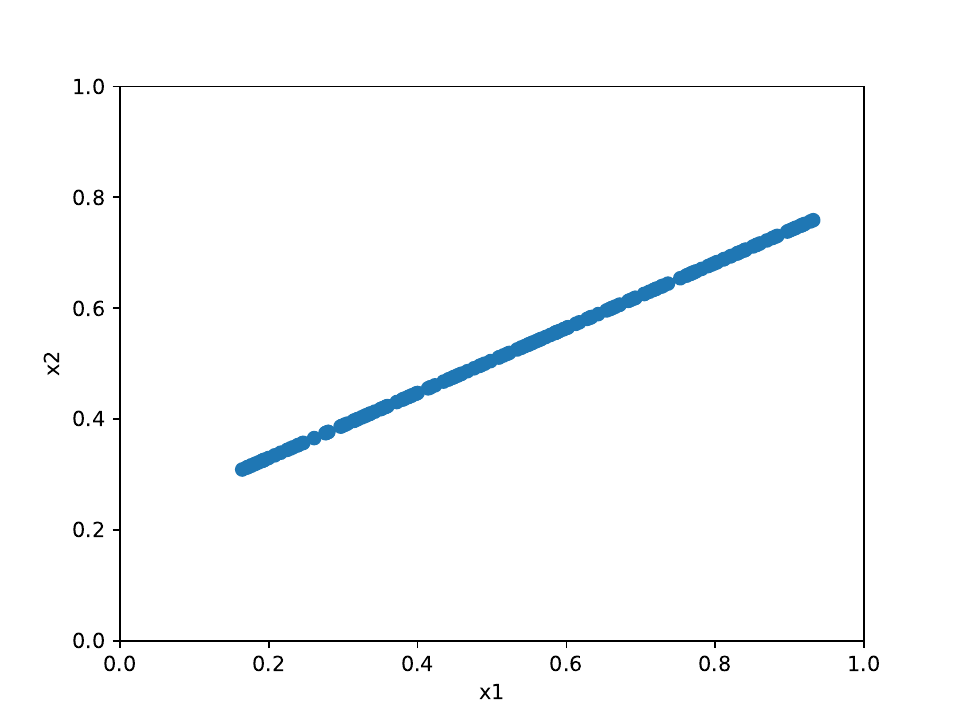}}
  \subfloat[Two clusters (C2)]{\includegraphics[width=1.4in]{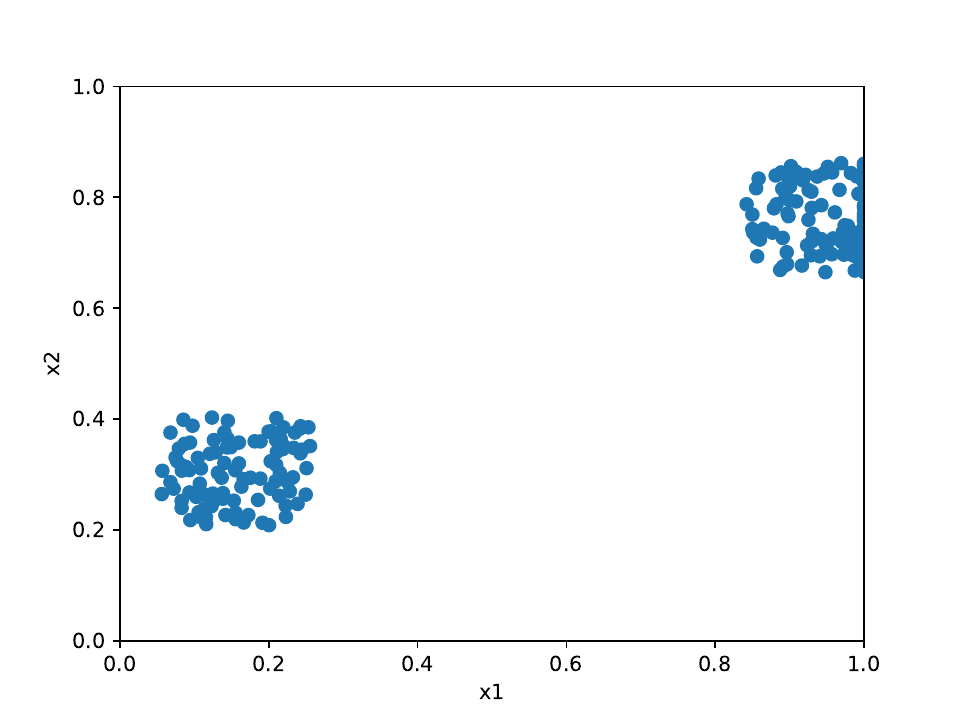}}
  \subfloat[Three clusters (C3)]{\includegraphics[width=1.4in]{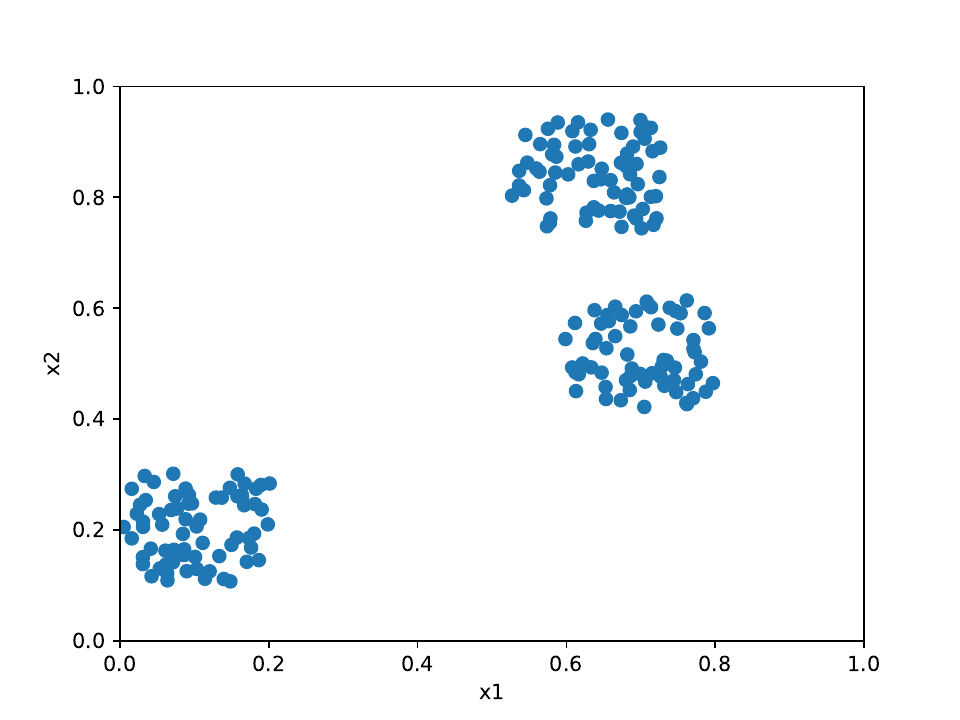}}
  \subfloat[Four clusters (C4)]{\includegraphics[width=1.4in]{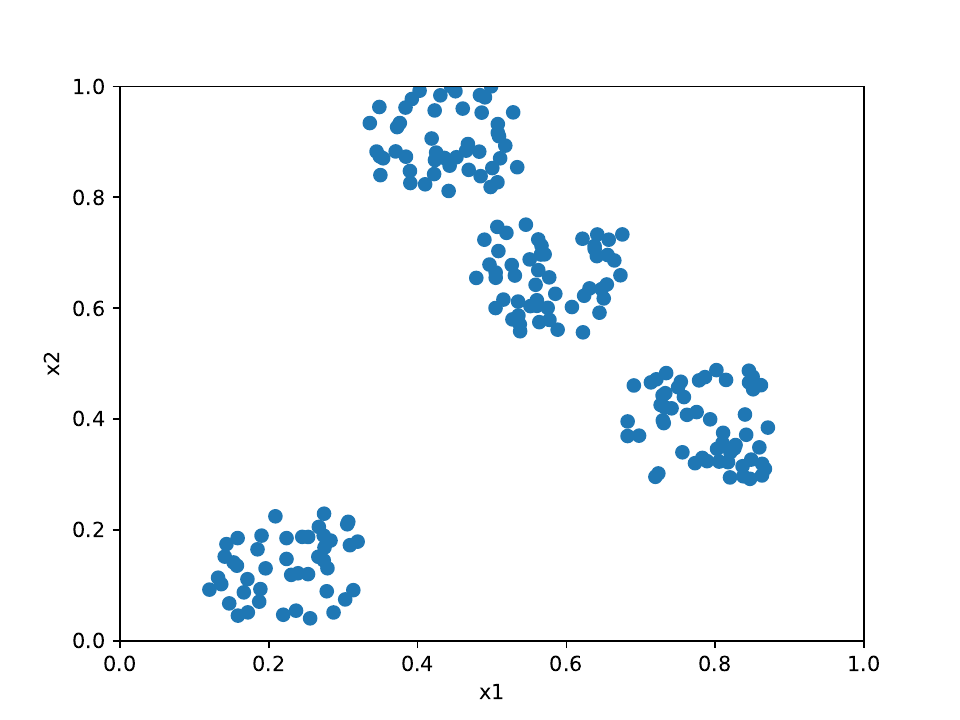}}
  \subfloat[Five clusters (C5)]{\includegraphics[width=1.4in]{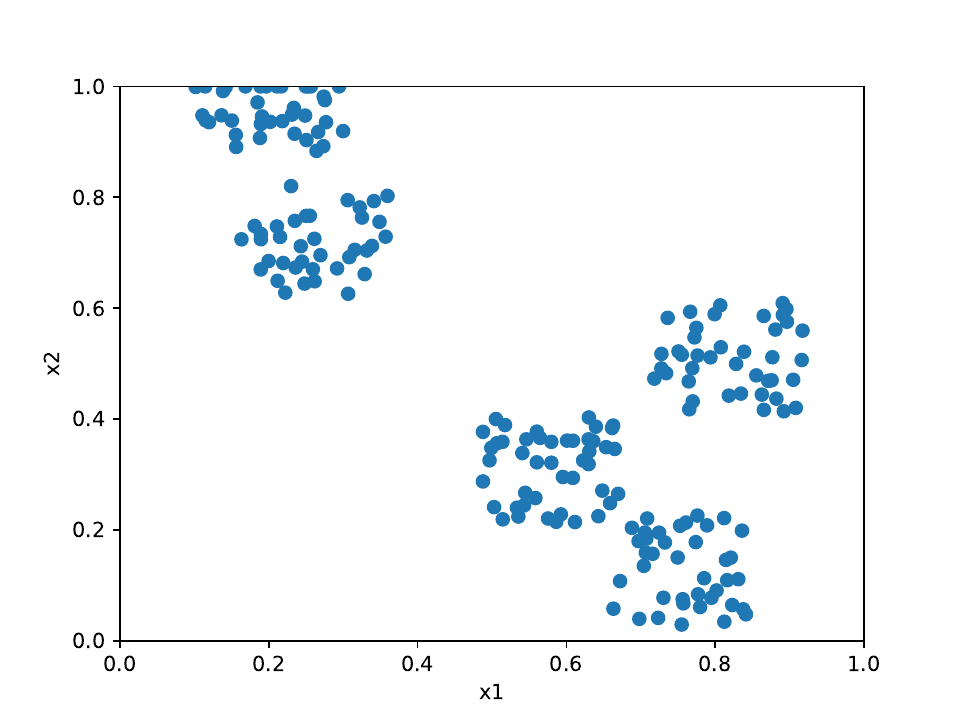}}
  \caption{The first two dimensions of different global optimum distributions for defining tasks, which varies from the range of optimal solutions (a)-(e) and the number of distributed clusters (f)-(j).}
  \label{fig: opt distribution}
  \vspace{-0.4cm}
\end{figure*}

To examine the adaptability of the EMT algorithms, we proposed two test suites with one constructed based on the CEC17MTOP benchmark \cite{da2017evolutionary} and the other one built upon the black-box optimization benchmark (BBOB) \cite{hansen2021coco}. For the first test suite, we use basic functions with configurable optimum from CEC17MTOP benchmark \cite{da2017evolutionary} and define 10 sets with different distributions of the global optimum for defining optimization tasks. The defined MTOP sets have various properties on global optimum distribution range and the number of distributed clusters in the solution space. The distribution of the first two dimensions of the task optimum of the ten sets is shown in Fig. \ref{fig: opt distribution}. For the second test suite, we use 24 synthetic functions from BBOB with covering diverse fitness landscapes to serve as the component task of the MTOP instance. Hence, the second test suite poses greater challenges to the adaptability of EMT algorithms. We constructed 15 MTOP sets for testing the adaptability and one MTOP set for learning the agent. For the details of the problem setting and parameter configuration, refer to Section S.III in the supplementary material \cite{wu2024learning}. 

\subsubsection{Compared Algorithms}
Recall the motivation is to improve the adaptability of the implicit EMT, our L2T framework should be compared with other implicit EMT algorithms to demonstrate its effectiveness. Moreover, to the best of our knowledge, up to now, there exist no algorithms also working on automatically adapting the KT process allowed for comparison. We adopt two types of EMT algorithms with DE and GA as base solvers which will be compared with MTDE-L2T and MTGA-L2T, respectively. By keeping the base solver the same for comparison, the experimental results can fairly reflect which KT design contributes to more search efficiency gain. Regarding the DE-based implicit EMT algorithms, the compared algorithms include AEMTO \cite{xu2021evolutionary}, MFDE \cite{feng2017empirical}, MKTDE \cite{li2021meta}, MTDE-AD \cite{wang2021solving}, MTDE-B \cite{jin2019study}. Regarding the GA-based implicit EMT algorithms, the compared algorithms include MFEA \cite{gupta2015multifactorial}, MFEA2 \cite{bali2019multifactorial}, Generalized MFEA (GMFEA) \cite{ding2017generalized}, MFEA with Adaptive KT (MFEA-AKT) \cite{zhou2020toward}, and MTEA-AD \cite{wang2021solving}. The reason for choosing these algorithms for comparison is that they share some similar structures with the KT action design in the L2T framework. For example, MFDE transfers differential vectors and MTDE-B transfers base vectors. In fact, MFDE is a special case of the agent with fixed output action $a_1=0.5,a_2=0,a_3=1$. Therefore, comparing these EMT algorithms can reveal whether our L2T framework is capable of discovering a better KT process than the human-designed KT process.

\subsubsection{Performance Metric}
Each MTOP instance set contains 100 MTOP instances constructed either by sampling the task optimum from the predefined distribution specified in Table S.I or by randomly selecting $(fid, sid)$ from the function ID set $\mathcal{F}$ and the seed ID set $\mathcal{S}$ as detailed in Table S.II in the supplementary material \cite{wu2024learning}. For each MTOP instance, we conduct 20 independent runs of each EMT algorithm, tracking fitness across generations. We conduct the Wilcoxon rank sum test with $\alpha=0.05$ to statistically compare the L2T-based EMT algorithm against baselines for each task, treating all tasks as equally important. Unlike prior work, we devise a more stringent pairwise comparison criterion as follows.
\begin{definition}
  Given an MTOP instance $\mathcal{T}=\{f_k(x)\}_{k=1}^{K}$, a target algorithm $\mathcal{A}_{t}$ and a baseline algorithm $\mathcal{A}_{b}$, with the set $\mathcal{Y}_{\mathcal{A},k}=\{f_k(x_{k,g,r}^*)\}_{r=1}^{R}$ denoting the best-found fitness at generation $g$ of task $f_k$ over $R$ independent runs on $\mathcal{T}$ by $\mathcal{A}$, $\mathcal{A}_{t}$ is said to outperform $\mathcal{A}_{b}$, denoted as $\mathcal{A}_{t} < \mathcal{A}_{b}$, if and only if $\mathcal{Y}_{\mathcal{A}_t,k} \preceq \mathcal{Y}_{\mathcal{A}_b,k},\forall k\in{1,...,K}$, where the $\preceq$ denotes the hypothesis test for examining whether the sample $\mathcal{Y}_{\mathcal{A}_t,k}$ is significantly better than or the same as $\mathcal{Y}_{\mathcal{A}_b,k}$. If the hypothesis test results are statistically same on all tasks, the comparison result between two algorithms is marked as a tie, denoted as $\mathcal{A}_{t} \approx  \mathcal{A}_{b}$. Otherwise, the EMT algorithm $\mathcal{A}_{t}$ is considered to be worse than $\mathcal{A}_{b}$, $\mathcal{A}_{t} >  \mathcal{A}_{b}$.
  \label{definition 1}
\end{definition}
This criterion implies that sacrificing the optimization performance on one task to achieve performance improvement on the other task is not encouraged for an EMT algorithm. That is, if $A_t$ outperforms $A_b$ on some tasks but worse on others, it's marked as a loss. We report the comparative outcome of the L2T-based and peer EMT algorithms over all instances as a `Win/Tie/Lose' (W/T/L) count.

\subsection{Effectiveness of the L2T Framework}
One desired property of the proposed L2T framework is that the agent for performing KT can be efficiently and flexibly learned and tested on the problem set of interest. To this end, we conduct experiments to learn 10 agents with DE as the base solver respectively on 10 problem sets based on CEC17MTOP. These 10 MTOP instance sets are with different distributions which are used in the illustrative experiment in Section II-C. It should be noted that in this experiment training MTOP instances and testing MTOP instances are drawn from the same distribution. Therefore, learning on diverse MTOP distributions can examine the adaptability of L2T to deal with complex MTOP instances with different levels of task similarities. The comparative results between MTDE-L2T and other implicit EMT algorithms are shown in Table S.IV. The results show that our proposed MTDE-L2T outcompetes peer implicit EMT algorithms on most of the MTOP sets with varying task optimum distributions. This indicates the effectiveness of the proposed L2T framework in finding and adapting a proper KT process for solving unseen MTOPs with complex distribution of interest.

\begin{figure}[!t]
  \vspace{-0.4cm}
  \centering
  \captionsetup[subfloat]{farskip=0pt,captionskip=0pt}
  \subfloat[Varying optimum range]{\includegraphics[width=1.8in]{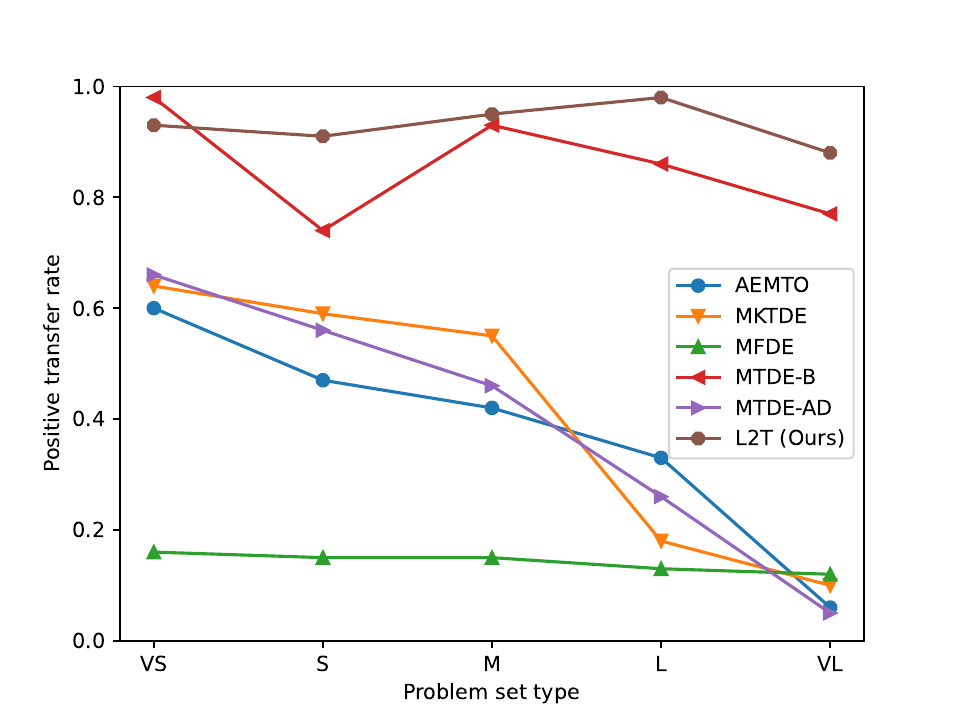} \label{fig: Varying optimum range}}
  \subfloat[Varying number of clusters]{\includegraphics[width=1.8in]{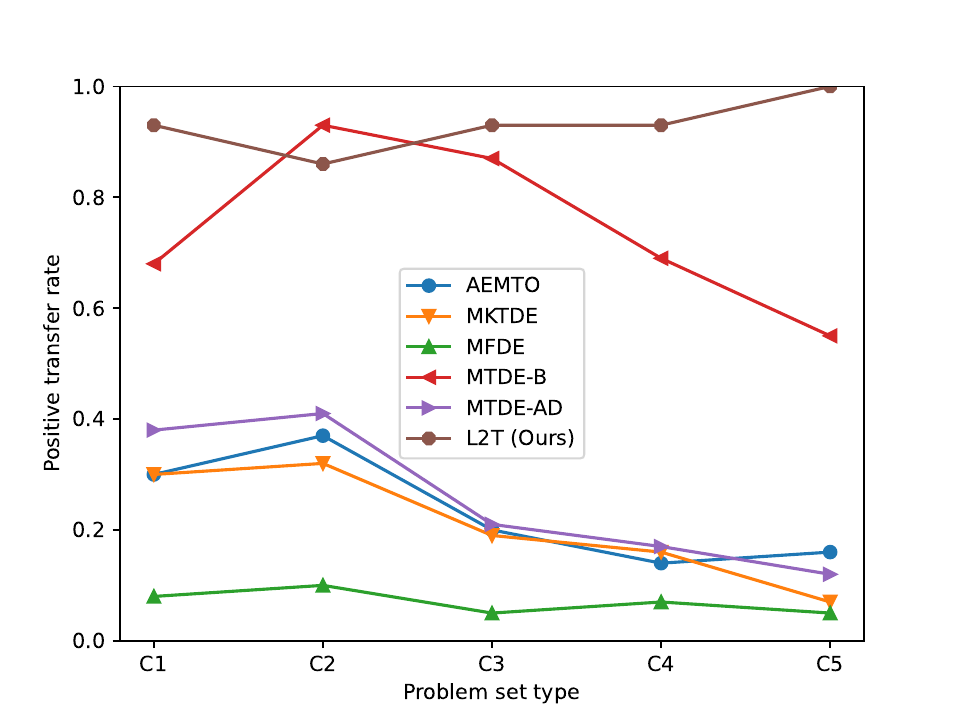} \label{fig: Varying number of clusters}}
  \caption{Positive transfer rates of implicit EMT algorithms on problem sets with varied task optimum range (a) and number of clusters (b).}
  \label{fig: positive transfer}
  \vspace{-0.4cm}
\end{figure}

Moreover, we investigate whether the agent learned by the L2T framework can bring more KT benefits with respect to its single-task counterpart, compared with other implicit EMT algorithms. To this end, we select implicit EMT algorithms using the same base solver DE on 10 MTOP sets of the first test suite and calculate the \emph{positive transfer rate} as
\begin{equation}
  \sum\nolimits_{i=1}^{N_{\mathcal{T}}}\mathbb{I}_{\mathcal{T}_i} (\mathcal{A}_{\rm{EMT}} < \mathcal{A}_{\rm{single}})/N_{\mathcal{T}}
\end{equation}
where $\mathcal{T}_i$ denotes the $i$-th MTOP instance, $\mathbb{I}(\cdot)$ denotes the indicator function, $N_{\mathcal{T}}=100$ denotes the number of MTOP instances drawn from a problem set, $\mathcal{A}_{\rm{EMT}}$ denotes a candidate EMT algorithm and $\mathcal{A}_{\rm{single}}$ denotes the single-task DE without performing any KT between tasks for solving the MTOP, and $<$ is the comparison criterion given in \textbf{Definition \ref{definition 1}}. The results reported in Fig. \ref{fig: positive transfer} show that for most implicit EMT algorithms, the ability to offer positive transfer results tends to deteriorate as the range MTOP distribution becomes broader (Fig. \ref{fig: positive transfer}\ref{sub@fig: Varying optimum range}) and the modality of the distribution becomes more complex (Fig. \ref{fig: positive transfer}\ref{sub@fig: Varying number of clusters}), while our proposed L2T-based EMT algorithm still yields high positive transfer rates. 

\subsection{Adaptability of the Learned Agent}
In this subsection, we first pre-trained two agents with DE and GA as base solvers on the $\text{BBOB}_{learn}$ and then examine the adaptability of MTDE-L2T and MTGA-L2T equipped with the learned agents on diverse MTOP sets based on BBOB (i.e., BBOB1-BBOB15). Since the tested MTOP sets involve more function classes with different characteristics and o.o.d. MTOP instances, these scenarios are considered more challenging than the MTOP sets based on CEC17MTOP. Moreover, we aim to obtain an adaptive and versatile agent as well as possible, we invest more computational budget with time steps $T=5e6$.
\subsubsection{Comparison with Implicit EMT Algorithms}
The comparative results of the proposed MTDE-L2T and MTGA-L2T and peer implicit EMT algorithms at generation=$G_{\rm{roll}}$ are shown in Table \ref{table: Comparative Results between the Proposed L2T and Implicit EMTO Algorithms at G_roll}. Specifically, we obtain three observations. First, the MTDE-L2T and MTGA-L2T generally outperform their competitors on MTOP sets (BBOB1-BBOB8) following nearly i.i.d. distribution with the training MTOP set $\text{BBOB}_{learn}$. These results indicate that the learned agent can successfully adapt to unseen and i.i.d MTOP instances to achieve competitive results. Second, the agent obtained by L2T can efficiently solve the o.o.d MTOP instance set (BBOB9-BBOB15) containing heterogeneous task functions that haven't been seen during the learning stage. These results indicate that the learned agent has the desired capability as we expect, i.e., adapting to a broad range of heterogeneous MTOP instances. Third, for the MTOP set using a single type of function but with different global optimum as tasks (e.g., BBOB3-BBOB8), the L2T-based algorithms also exhibit performance superiority over the compared EMT algorithm. This indicates that the learned agent also learns the effective skill of performing KT between tasks having homogeneous or highly similar function landscapes but with rather different optimums. As discussed in \cite{wu2023transferable}, the ability to utilize the shift invariance between tasks to speed up the search is also a desired property of KT. Moreover, it should be highlighted that MTDE-L2T with the learned agent generally outperforms the manually designed MFDE and MTDE-B whose KT behaviors actually lie in our designed action space. These results demonstrate the promise of our L2T framework in automatically discovering efficient KT processes for handling MTOPs with broader distributions than the traditionally designed KT process driven by human expertise. Next, we report the comparative results between L2T-based and other implicit EMT algorithms at generation=$G_{\max}$ in Table S.V in the supplementary material \cite{wu2024learning}. Note that since the optimization steps that can be experienced by the agent during the learning stage is $G_{\rm{roll}}=100<G_{\max}$, the performance results at generation=$G_{\max}$ actually reflect the adaptability of the agents to generalize to longer optimization horizon (i.e., the evolution stage that has not seen before). From the table, we find that the performance of L2T-based algorithms at generation=$G_{\max}$ has a drop compared with that at generation=$G_{\rm{roll}}$. However, the L2T-based EMT algorithms still offer performance advantages over their competitors, demonstrating the capability of the learned agent to provide long-term efficiency benefits in EMT.

\begin{table*}[htbp]\scriptsize
  \setlength{\abovecaptionskip}{0cm}
  \setlength{\belowcaptionskip}{-0.2cm}
  \centering
  \setlength{\tabcolsep}{4pt} 
  \caption{Comparative Results between the Proposed L2T-based and Other Implicit EMT Algorithms at Generation=$G_{\rm{roll}}$}
  \begin{tabular}{ccccccccccc}
    \toprule
    \multirow{2}[2]{*}{\textbf{Problem}} & \multicolumn{5}{c}{\textbf{MTDE-L2T vs}} & \multicolumn{5}{c}{\textbf{MTGA-L2T vs}} \\
    \cmidrule(r){2-6} \cmidrule(l){7-11}  & \textbf{AEMTO} & \textbf{MFDE} & \textbf{MKTDE} & \textbf{MTDE-AD} & \textbf{MTDE-B} & \textbf{GMFEA} & \textbf{MFEA} & \textbf{MFEA-AKT} & \textbf{MFEA2} & \textbf{MTEA-AD} \\
    \midrule
    BBOB1 & 56/43/1($+$) & 63/30/7($+$) & 88/10/2($+$) & 53/46/1($+$) & 36/59/5($+$) & 83/10/7($+$) & 82/13/5($+$) & 84/12/4($+$) & 76/19/5($+$) & 76/12/12($+$) \\
    BBOB2 & 68/30/2($+$) & 61/32/7($+$) & 85/11/4($+$) & 57/38/5($+$) & 43/50/7($+$) & 84/9/($+$) & 85/11/4($+$) & 77/18/5($+$) & 78/19/3($+$) & 78/10/12($+$) \\
    BBOB3 & 100/0/0($+$) & 14/84/2($+$) & 5/86/9($-$) & 100/0/0($+$) & 11/86/3($+$) & 100/0/0($+$) & 100/0/0($+$) & 100/0/0($+$) & 100/0/0($+$) & 100/0/0($+$) \\
    BBOB4 & 30/69/1($+$) & 12/84/4($+$) & 98/2/0($+$) & 7/91/2($+$) & 1/96/3($-$) & 100/0/0($+$) & 100/0/0($+$) & 100/0/0($+$) & 95/5/0($+$) & 100/0/0($+$) \\
    BBOB5 & 8/89/3($+$) & 5/88/7($-$) & 4/87/9($-$) & 7/93/0($+$) & 3/91/6($-$) & 91/7/2($+$) & 89/9/2($+$) & 86/11/3($+$) & 92/7/1($+$) & 86/11/3($+$) \\
    BBOB6 & 25/74/1($+$) & 40/59/1($+$) & 77/23/0($+$) & 12/83/5($+$) & 3/85/12($-$) & 12/71/17($-$) & 8/70/22($-$) & 10/71/19($-$) & 24/71/5($+$) & 7/56/37($-$) \\
    BBOB7 & 9/86/5($+$) & 8/89/3($+$) & 7/91/2($+$) & 6/90/4($+$) & 5/84/11($-$) & 18/77/5($+$) & 20/75/5($+$) & 20/77/3($+$) & 24/74/2($+$) & 1/79/20($-$) \\
    BBOB8 & 34/65/1($+$) & 23/74/3($+$) & 87/13/0($+$) & 13/85/2($+$) & 5/88/7($-$) & 98/2/0($+$) & 98/2/0($+$) & 88/12/0($+$) & 83/17/0($+$) & 100/0/0($+$) \\
    BBOB9 & 42/57/1($+$) & 51/35/14($+$) & 67/23/10($+$) & 45/52/3($+$) & 18/66/16($+$) & 59/33/8($+$) & 60/29/11($+$) & 60/28/12($+$) & 64/29/7($+$) & 63/21/16($+$) \\
    BBOB10 & 51/33/16($+$) & 55/36/9($+$) & 62/29/9($+$) & 54/35/11($+$) & 27/48/25($+$) & 78/16/6($+$) & 78/18/4($+$) & 79/18/3($+$) & 79/18/3($+$) & 72/21/7($+$) \\
    BBOB11 & 96/0/4($+$) & 27/70/3($+$) & 13/85/2($+$) & 96/0/4($+$) & 16/82/2($+$) & 100/0/0($+$) & 100/0/0($+$) & 100/0/0($+$) & 100/0/0($+$) & 99/0/1($+$) \\
    BBOB12 & 49/51/0($+$) & 1/30/69($-$) & 1/30/69($-$) & 45/53/2($+$) & 15/80/5($+$) & 76/22/2($+$) & 75/24/1($+$) & 83/16/1($+$) & 82/18/0($+$) & 86/13/1($+$) \\
    BBOB13 & 88/9/3($+$) & 28/68/4($+$) & 12/78/10($+$) & 94/5/1($+$) & 12/81/7($+$) & 99/0/1($+$) & 98/0/2($+$) & 99/0/1($+$) & 99/0/1($+$) & 98/0/2($+$) \\
    BBOB14 & 11/86/3($+$) & 10/87/3($+$) & 10/88/2($+$) & 6/89/5($+$) & 8/89/3($+$) & 26/74/0($+$) & 24/75/1($+$) & 30/70/0($+$) & 42/57/1($+$) & 0/81/19($-$) \\
    BBOB15 & 14/82/4($+$) & 15/79/6($+$) & 23/73/4($+$) & 18/78/4($+$)& 7/84/9($-$) & 25/51/24($+$) & 24/54/22($+$) & 20/56/24($-$) & 23/59/18($+$) & 23/54/23($=$) \\
    \bottomrule
  \end{tabular}%
  \label{table: Comparative Results between the Proposed L2T and Implicit EMTO Algorithms at G_roll}
\end{table*}%

\subsubsection{Comparison with Explicit EMT Algorithms}
Besides the comparison with implicit EMT algorithms, we investigate how well the L2T-based algorithms perform when compared with state-of-the-art explicit EMT algorithms that involve an explicit learning process to perform KT. We choose MTDE with Explicit Autoencoding (MTDE-EA) \cite{feng2018evolutionary} and Affine Transformation-based MFEA (ATMFEA) \cite{xue2020affine} as the DE-based and GA-based explicit EMT algorithms, respectively for comparison. The comparative results at different generations $G_{\rm{roll}}, G_{\max}$ are reported in Table S.VI in the supplementary material \cite{wu2024learning}. The results show that the performances of the proposed L2T-based algorithms are generally better than the compared explicit EMT algorithms in terms of the adaptability to various MTOPs with different distributions, further illustrating the preeminence of the L2T framework.

\begin{figure*}[!t]
  \centering
  \captionsetup[subfloat]{farskip=-2pt,captionskip=0pt}
  \subfloat[BBOB9]{\includegraphics[width=1.4in]{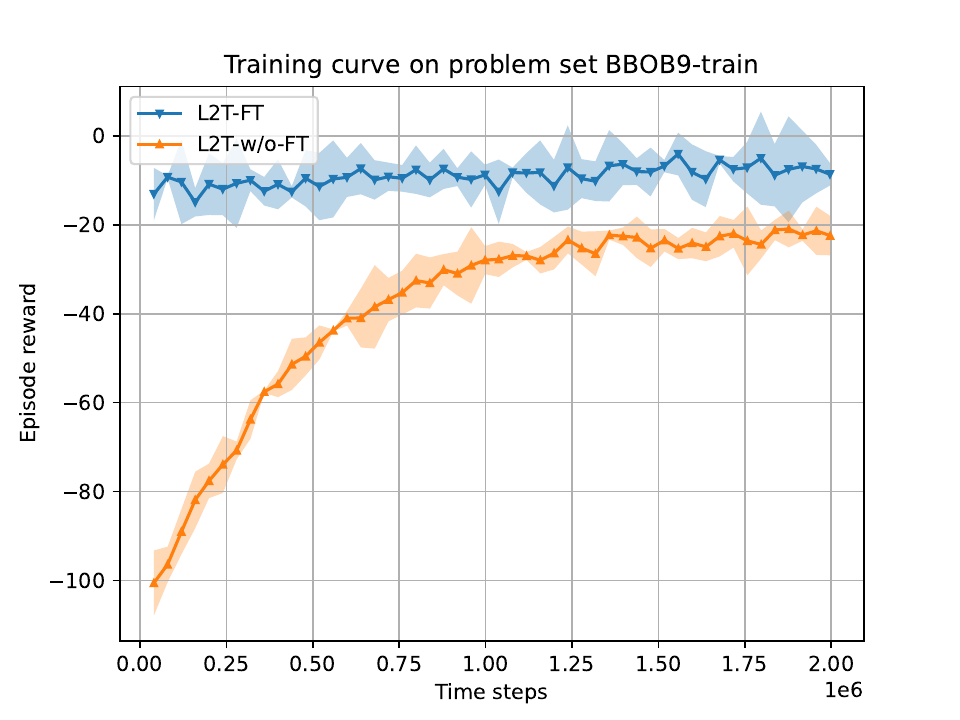}}
  \subfloat[BBOB10]{\includegraphics[width=1.4in]{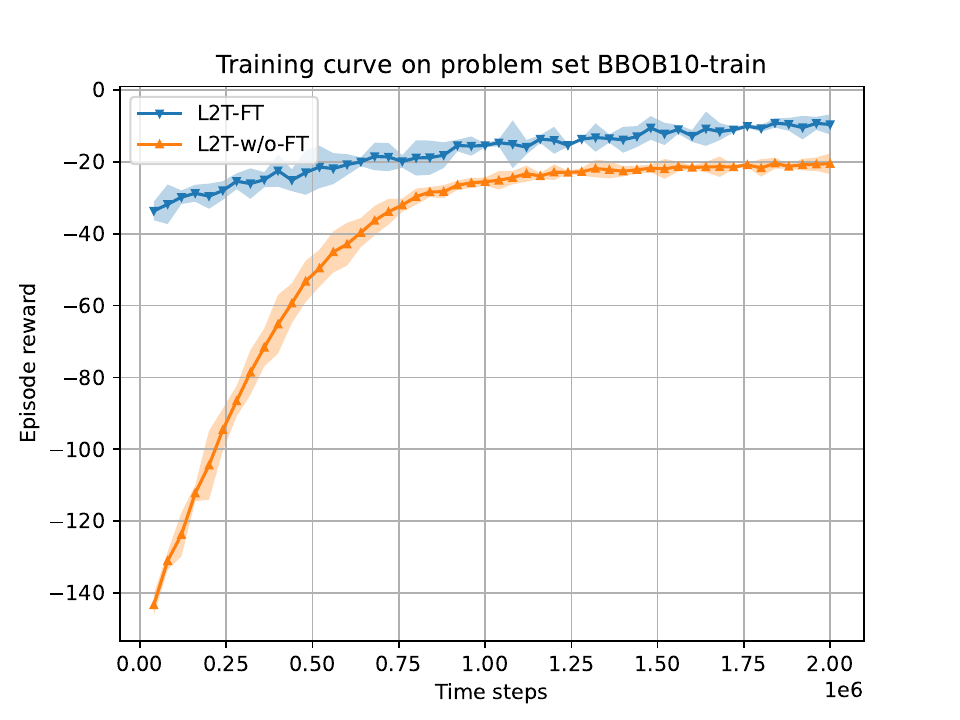}}
  \subfloat[VL]{\includegraphics[width=1.4in]{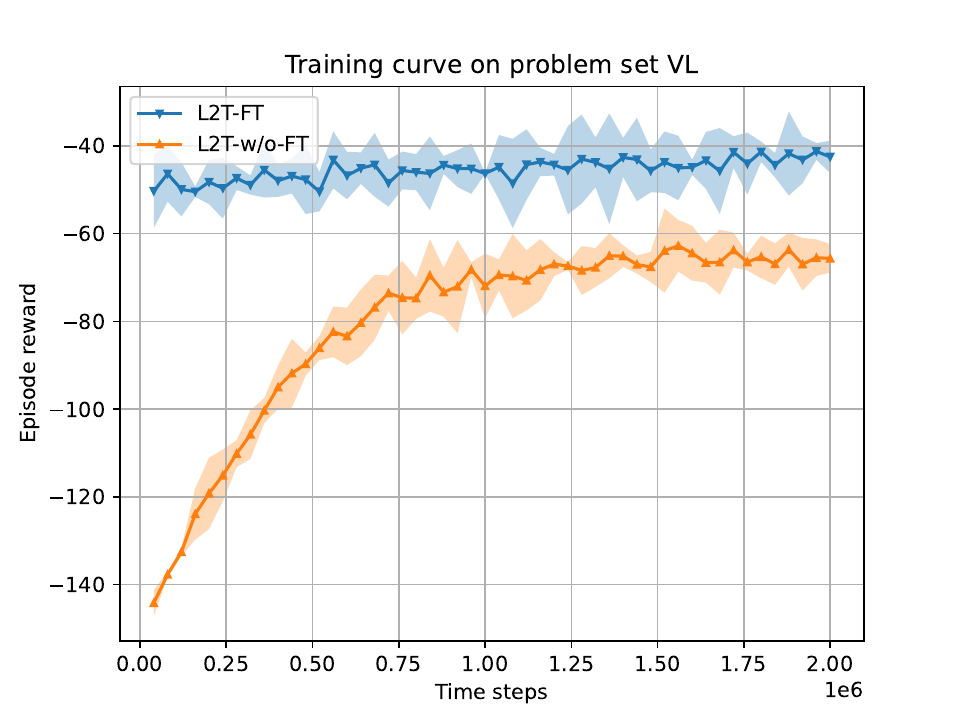}}
  \subfloat[C1]{\includegraphics[width=1.4in]{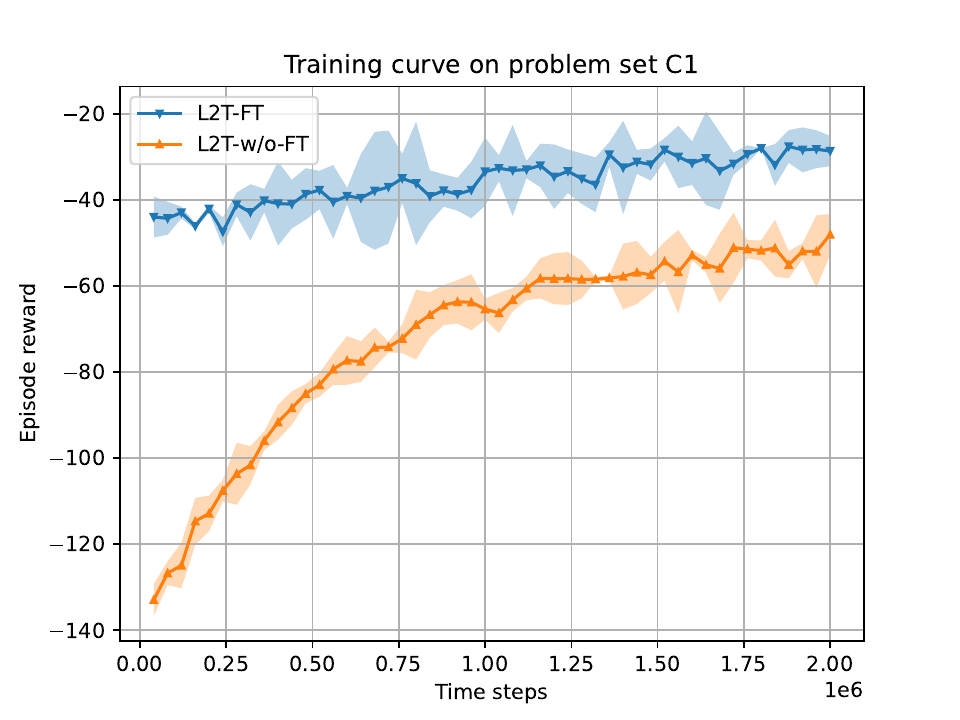}}
  \subfloat[C5]{\includegraphics[width=1.4in]{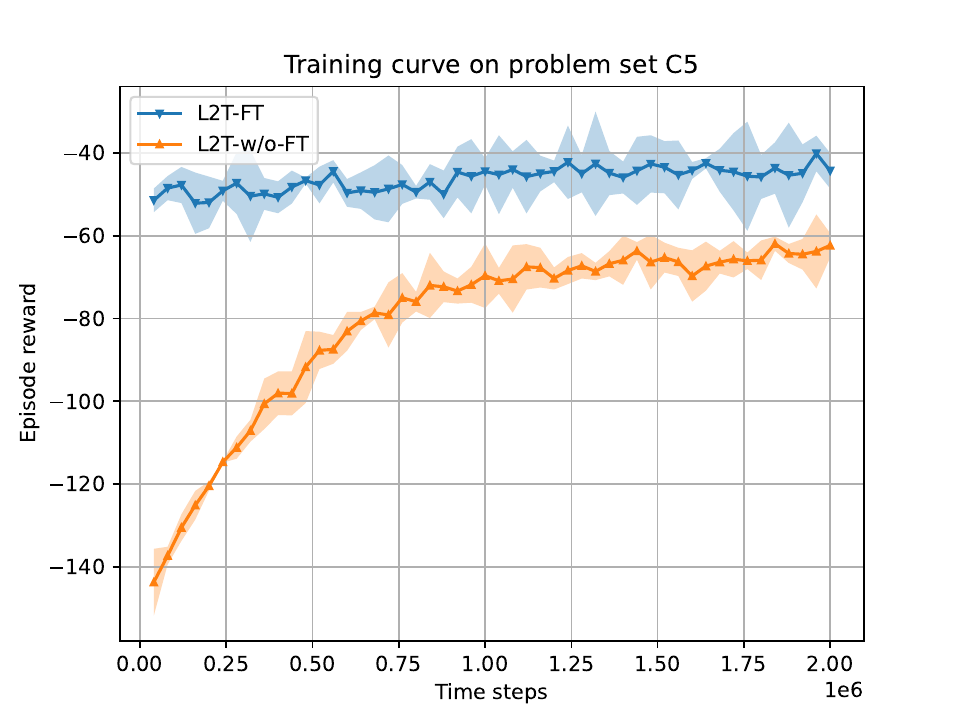}}
  \caption{The training performance of agents by fine-tuning (L2T-FT) and retraining from scratch (L2T-w/o-FT) on new problem sets with significantly different task distributions.}
  \label{fig: fine-tuning and retraining performance}
  \vspace{-0.4cm} %
\end{figure*}

\subsection{Transferability of the Learned Agent}
We are interested in the reusability and transferability of the agent pre-trained on $\text{BBOB}_{learn}$ to solve new MTOP sets. We conduct experiments on 10 problem sets based on CEC19MTOP plus BBOB9 and BBOB10 by fine-tuning the agent that is pre-trained on $\text{BBOB}_{learn}$. The agents in Section IV-B are regarded as learning from scratch since it does not reuse well-trained agents from other problem sets. The resultant algorithms with the agent learned from scratch and learned by fine-tuning are termed MTDE-L2T without fine-tuning (MTDE-L2T-w/o-FT) and MTDE-L2T by fine-tuning (MTDE-L2T-FT), respectively. For MTDE-L2T-w/o-FT, two agents are also learned from scratch on BBOB9 and BBOB10 respectively. The comparative results between MTDE-L2T-w/o-FT and MTDE-L2T-FT and DE-based EMT algorithms are shown in Table S.VII within the supplementary material \cite{wu2024learning}. We gain two observations from Table S.VII. First, the learned MTDE-L2T generally outperforms the peer EMT algorithms, indicating the applicability of the proposed L2T framework on different problem sets of interest. Second, MTDE-L2T-FT yields better results than the MTDE-L2T-w/o-FT, illustrating the transferability of the pre-trained agent on $\text{BBOB}_{learn}$ to new problem sets. Furthermore, the performance along the learning process of fine-tuning (L2T-FT) and learning from scratch (L2T-w/o-FT) on five problem sets are shown in Fig. \ref{fig: fine-tuning and retraining performance}. It can be seen that on most problem sets, L2T-FT converges faster and better than L2T-w/o-FT, indicating that L2T-FT with the pre-trained agent normally requires less computational resources to achieve high performance. These results further demonstrate the transferability of the proposed L2T framework.

\subsection{Component Analysis}
We first analyze the effectiveness of the learning process of the L2T framework. To this end, several agents with fixed action and random action are formulated. Specifically, we use `MTDE-f($x$,$y$,$z$)' to denote the EMT algorithm with an agent that simply uses a fixed action $a_1=x,a_2=y,a_3=z$ for each task along the search process. Moreover, we use `MTDE-r' to denote the agent that randomly draws a sample from $[0,1]^3$ for each task each time as the action to perform KT. Furthermore, the learned agent should be compared with the single-task DE (STDE) that does not transfer any cross-task information, to reveal that the performance gain is actually caused by the effective utilization of transferable knowledge between tasks. The comparative results between the MTDE-L2T and the above-mentioned algorithms are given in Table S.VIII. The results show that MTDE-L2T yields significantly better results than the compared algorithms. These results indicate that (1) the learning algorithm PPO used in the L2T framework helps find well-performing agents than the trivial agents with fixed and random actions, and (2) the L2T framework successfully learns to identify and harness useful cross-task knowledge to improve search efficiency.

Next, we conduct ablation studies to verify the designed components in the L2T framework indeed contribute to the outcomes of good performance. In particular, the plausibility and the effectiveness of the action space design, extracted features as state space, and the reward design with two proposed additional reward terms should be examined. Regarding the effects of action space, we respectively remove each action from the designed three-dimensional action space to learn the agent. Then we formulate three variants, namely, `L2T-w/o-$a_1$', `L2T-w/o-$a_2$', and `L2T-w/o-$a_3$'. Accordingly, the number of output units of the actor network is reduced to $2\cdot K$, while the default settings of the removed actions are set as $a_1=0.5,a_2=0,a_3=0$, respectively. Regarding the effects of the observation space, we respectively remove the common features $O_c$ and task-specific features $O_t$ from the original designed state space, to formulate two variants, i.e., `L2T-w/o-$O_c$' and `L2T-w/o-$O_t$'. The modification on the actor networks of `L2T-w/o-$O_c$' and `L2T-w/o-$O_t$' is similar to that of `L2T-w/o-$a_1$' except that the number of input units will be reduced accordingly. To verify the overall effects of the devised feature extraction, we formulate a variant named `L2T-w/o-$FE$' that uses the full state of the population and fitness of all the tasks discussed in Section III-B-3). Regarding the effects of the reward design, we formulate two variants called `L2T-w/o-$r_{conv}$' and `L2T-w/o-$r_{KT}$', where the reward term $r_{conv}$ and $r_{KT}$ are removed from them, respectively. It should be noted that each of the formulated variants is independently learned on distinct problem sets and subsequently tested on the corresponding problem sets, consistent with the approach taken by MTDE-L2T.

The results between the MTDE-L2T and the above-mentioned variants are reported in Table S.IX in the supplementary material \cite{wu2024learning}. It can be observed that the proposed MTDE-L2T obtains more `$+$' than `$-$' when compared with L2T-w/o-$a_1$. Notably, MTDE-L2T is superior on i.i.d problem sets that have confirmed similarity with the learning problem set. This result indicates that the transfer intensity needs to be adaptively adjusted according to the evolution status and our MTDE-L2T can automatically learn useful skills in deciding transfer intensity. When removing $a_2$ and $a_3$ from our design, we can also witness a degradation in the performance, which validates the effectiveness of the proposed action design in Eq. (\ref{eq: sample_mutation_vector}). Moreover, MTDE-L2T yields generally better results when comparing the variants with removed state features, i.e., L2T-w/o-$O_c$, L2T-w/o-$O_t$, and L2T-w/o-$FE$. This indicates that the proposed state feature extraction is beneficial by incorporating more inter-task and intra-task information to facilitate wise decision-making. Finally, MTDE-L2T's optimization performances surpass L2T-w/o-$r_{conv}$ and L2T-w/o-$r_{KT}$, demonstrating the usefulness of the two proposed reward terms in assisting the agent to learn faster. One more notable observation is that the proposed reward term $r_{KT}$ seems to contribute more to delivering high performance than $r_{conv}$.

\subsection{Agent Behavior Analysis}
\begin{figure*}[htbp]
  \setlength{\abovecaptionskip}{0.cm}
  \setlength{\belowcaptionskip}{-0.2cm}
  \captionsetup[subfloat]{farskip=0pt,captionskip=0pt}
  \centering

  \subfloat[$O_t$ for $f_1$ on MTOP15]{\includegraphics[width=1.4in]{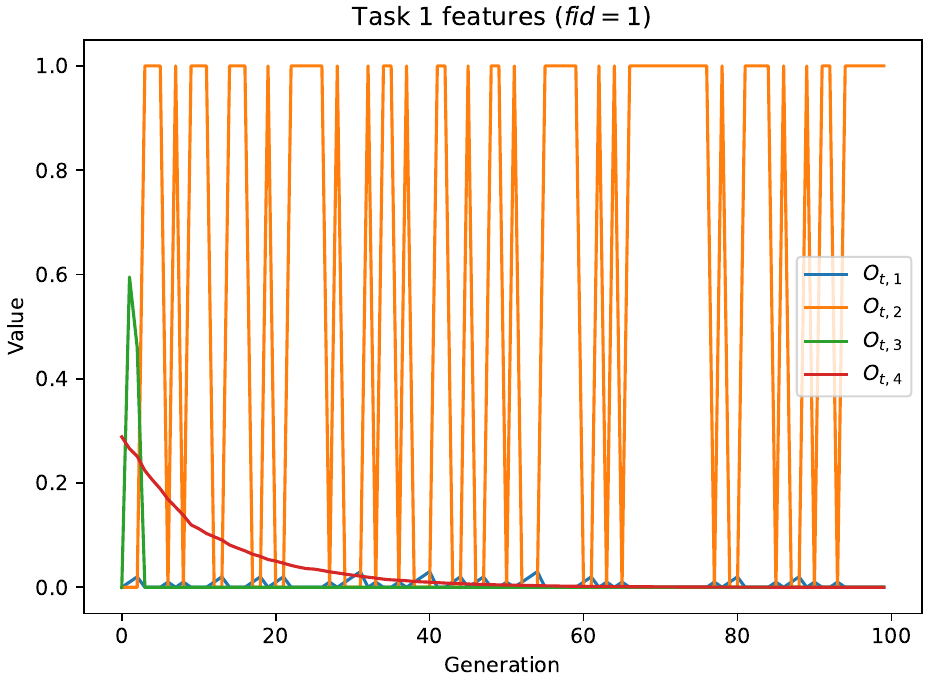}\label{fig: b}}
  \subfloat[$O_t$ for $f_2$ on MTOP15]{\includegraphics[width=1.4in]{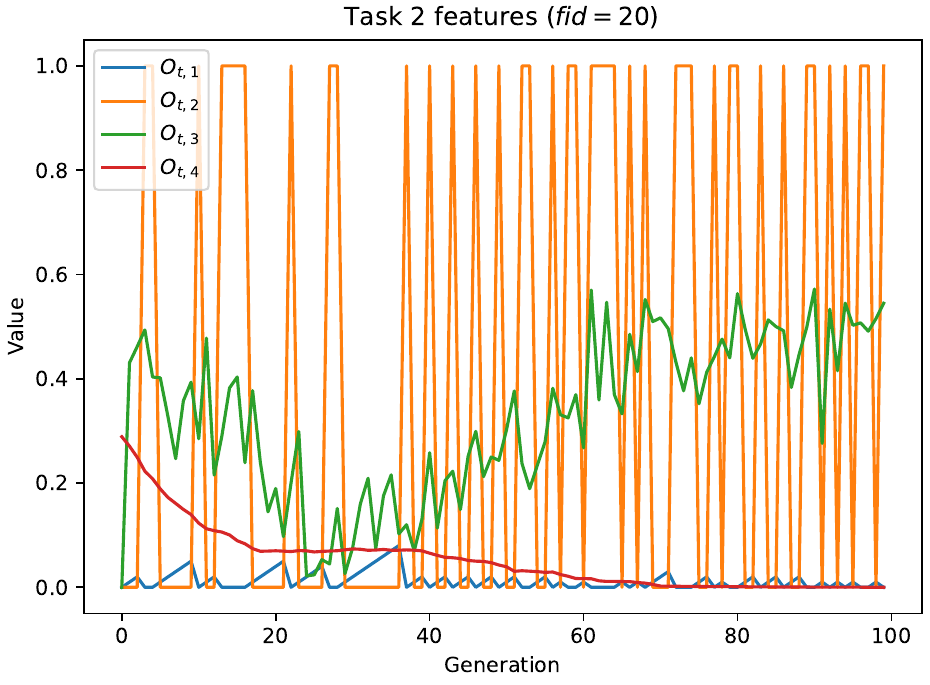}\label{fig: c}}
  \subfloat[$a$ for $f_1$ on MTOP15]{\includegraphics[width=1.4in]{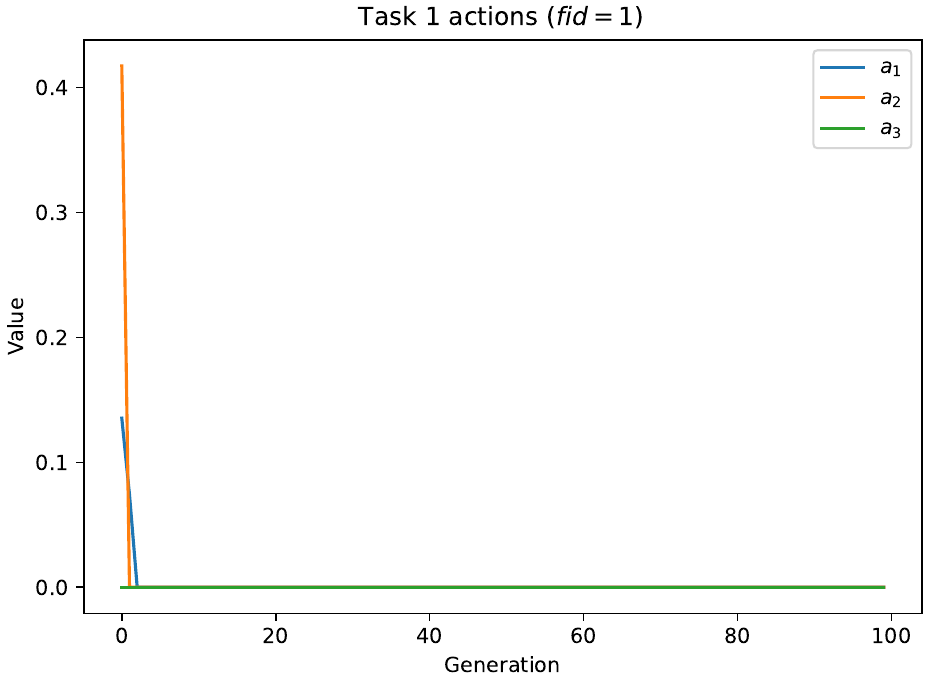}\label{fig: d}}
  \subfloat[$a$ for $f_2$ on MTOP15]{\includegraphics[width=1.4in]{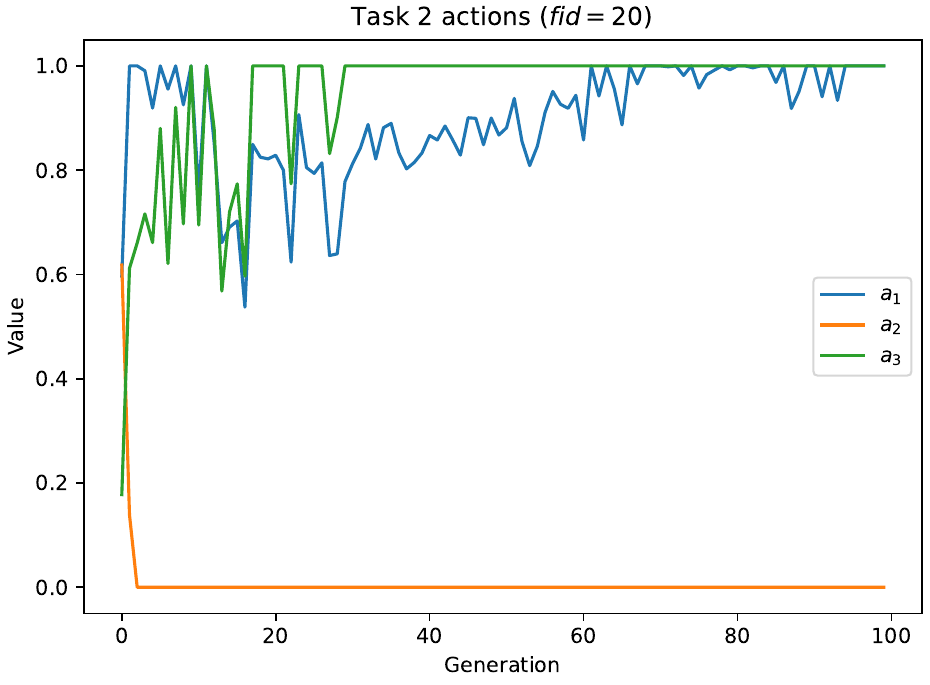}\label{fig: e}}

  \subfloat[$O_t$ for $f_1$ on MTOP23]{\includegraphics[width=1.4in]{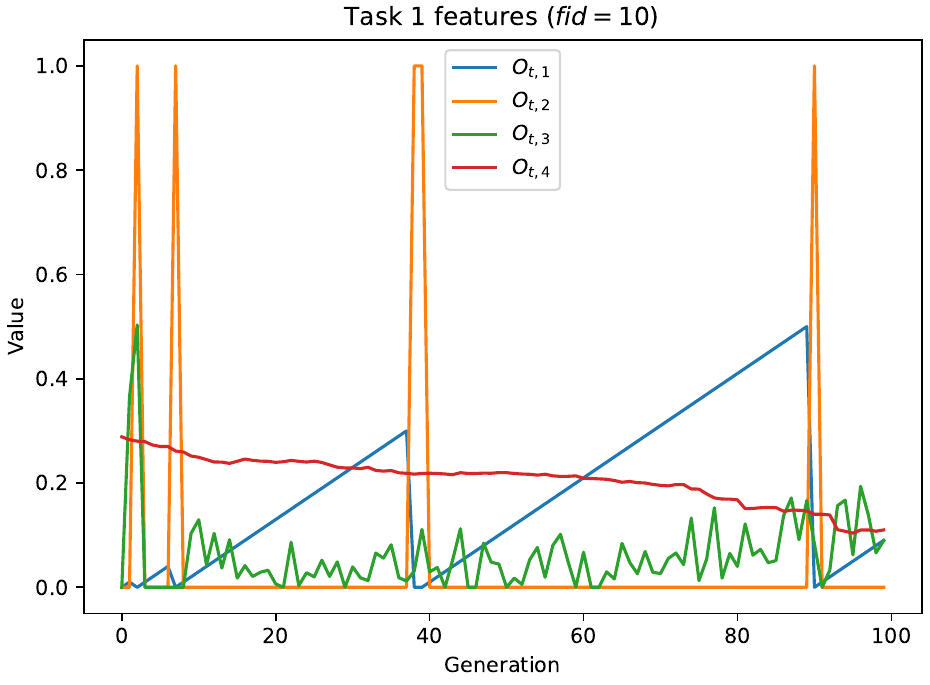}\label{fig: g}}
  \subfloat[$O_t$ for $f_2$ on MTOP23]{\includegraphics[width=1.4in]{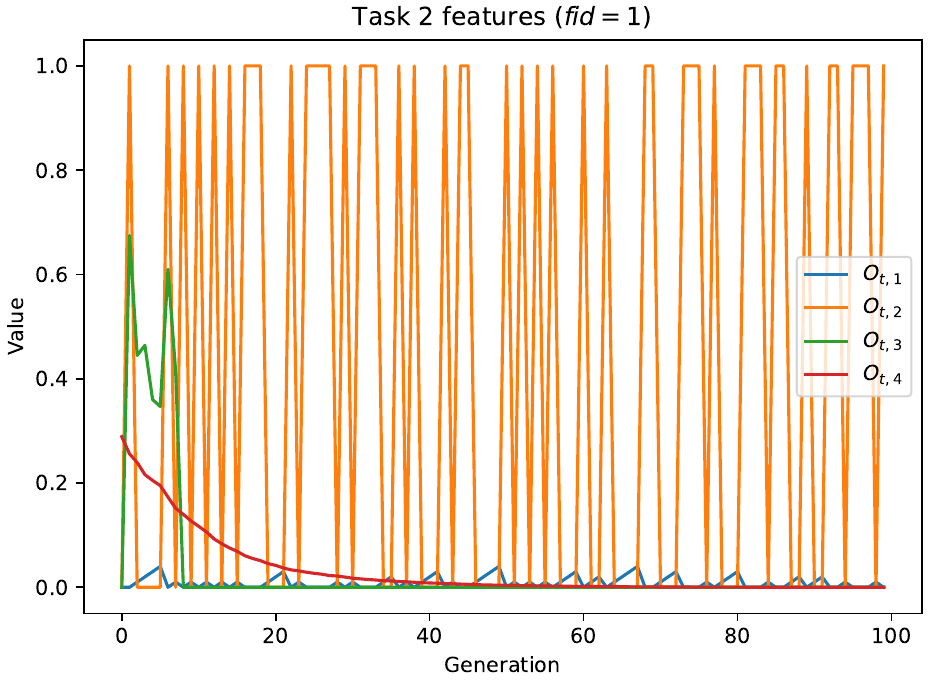}\label{fig: h}}
  \subfloat[$a$ for $f_1$ on MTOP23]{\includegraphics[width=1.4in]{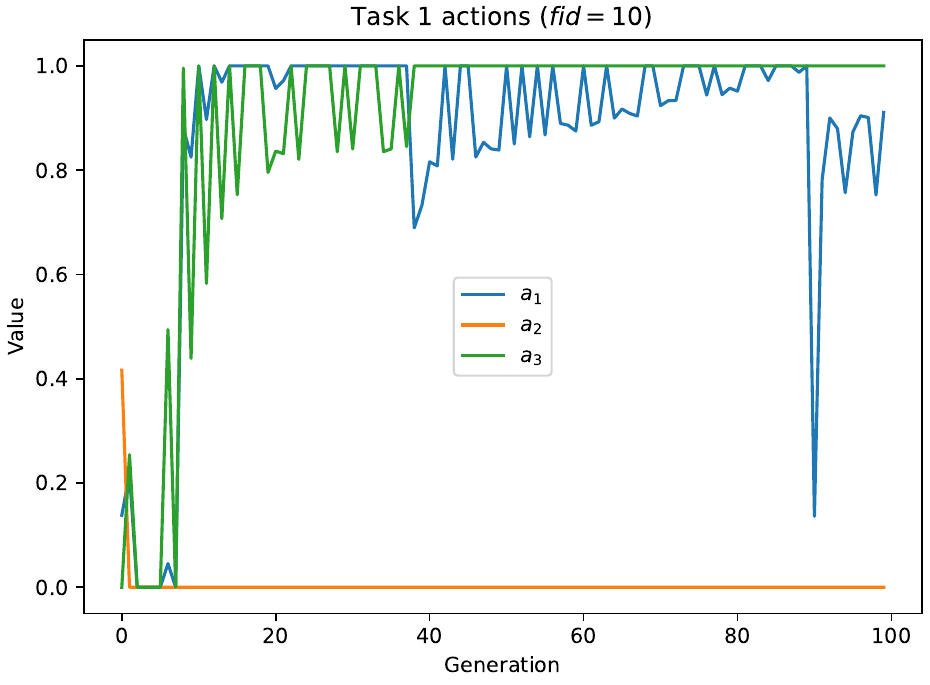}\label{fig: i}}
  \subfloat[$a$ for $f_2$ on MTOP23]{\includegraphics[width=1.4in]{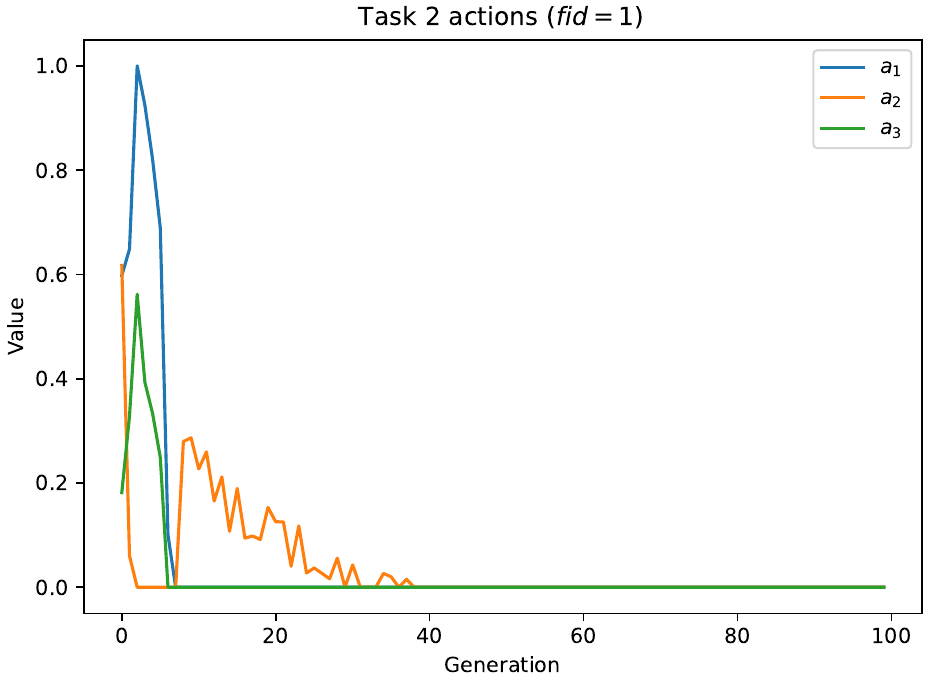}\label{fig: j}}
  \caption{The actions by the learned agent when solving two MTOP instances, MTOP15 (a)-(e) and MTOP23 (e)-(j) in the BBOB1 problem set.}
  \label{fig: behaviors}
  \vspace{-0.4cm} %
\end{figure*}

The pattern of making KT decisions of the learned agent is analyzed in this subsection. The observed states and the actions taken by the agent along the search process on two MTOP instances of the BBOB1 problem set are plotted in Fig. \ref{fig: behaviors}. The MTOP15 and MTOP23 are selected as representative cases to analyze the behaviors. MTOP15 contains an unimodal function $f_1$ as task 1 and a highly multimodal function $F_{20}$ as task 2. MTOP23 contains two unimodal functions $f_1,f_{10}$ as tasks with differences in the conditioning number of the landscape. It can be seen in Fig. \ref{fig: behaviors} that when solving MTOP15, the agent learns the behavior of performing unidirectional KT from task 1 to task 2 with ($a_1\geq 0$) while keeping the task 2 evolve without KT ($a_1\approx 0$). This learned strategy can mainly be translated as using easy task (task 1) to help difficult task (task 2) search efficiently, which has been reported to be useful in literature \cite{ding2017generalized}. In the case of MTOP15 having distant task optimum, the learned agent favors transferring differential vector ($a_3> 0$) rather than base vectors ($a_2\approx 0$), the plausibility of this KT behavior has also been supported by \cite{li2021meta}. For MTOP23 with two unimodal functions, the learned agent also manages to gradually adapt the KT action to improve search performance. First, we observe that the agent can dynamically adjust transfer intensity ($a_1$) and the way of performing KT ($a_2,a_3$) according to the evolution state (Fig. \ref{fig: behaviors}\ref{sub@fig: i}). Second, the agent learns to give a high base vector transfer ratio ($a_2 > 0$) in the early stage of the optimization process (Fig. \ref{fig: behaviors}\ref{sub@fig: j}). This behavior is reasonable since populations are highly overlapped in the beginning generations, due to the uniform initialization in the whole search space. Then base vector transfer between tasks in the early stage can help locate promising regions faster. As the search proceeds, the populations gradually converge to different optimum, the base vector transfer rate decreases (Fig. \ref{fig: behaviors}\ref{sub@fig: j}) since direct solution transfer is not useful under this condition. 

\subsection{Parameter Sensitivity Analysis}
The key parameters of the proposed L2T include $b_1, b_2$ in the proposed reward design. Note that $b_3$ is directly related to the final performance and is suggested to be fixed. The component analysis in Section IV-E has suggested that setting $b_1=0$ or $b_2=0$ would lead to deteriorated results. The settings of $b_1,b_2$ reflect the balance between the convergence gain term $r_{conv}$ and the KT gain term $r_{KT}$. Therefore, we mainly investigate the effect of relative setting of $b_1,b_2$ by keeping $b_1=1$ unchanged and varying the parameter $b_2\in\{0.1,0.5,1,5,10,50,100\}$ to investigate its sensitivity. Specifically, we learned the agent under different settings of $b_2$ and tested the performance of the M2DE-L2T on the 15 BBOB-based problem sets. We use the mean standard score (MSS) \cite{da2017evolutionary} as the metric to quantitatively compare the performance of different settings. Then the average ranking computed based on the MSS will be reported. The investigation results on parameter $b_2$ at generation=$G_{\rm{roll}}$ and generation=$G_{\max}$ are given in Table S.X and Table S.XI in the supplementary material \cite{wu2024learning}. From the two tables, we can observe that for the BBOB problem set $b_2=10$ and $b_2=5$ yield the best performance for the optimization horizon of $G_{\rm{roll}}$ and $G_{\max}$, respectively. When adopting the L2T framework to solve new MTOP sets, the grid search can be conducted on the parameter $b_2$ to fulfill its capability.

\subsection{Real-World Application}
In this subsection, we investigate the performance of the proposed L2T framework in solving real-world application MTOPs. To this end, we employ the hyperparameter optimization (HPO) task for machine learning methods \cite{klein2019meta} as the testbed. Specifically, an HPO task can be represented by the tuple $(\mathcal{H},\mathcal{D},\mathcal{L})$ with a hyperparameter space $\mathcal{H}$, a dataset $\mathcal{D}$, and a learning algorithm $\mathcal{L}$. The hyperparameter space dimensionality is denoted as $D_{\mathcal{H}}$. Especially, we adopt three task sets for optimizing the hyperparameters of machine learning algorithms including support vector machine (SVM) with dimension $D_{\mathcal{H}}=2$, XGBoost with dimension $D_{\mathcal{H}}=8$, fully-connected neural networks (FCNet) with dimension $D_{\mathcal{H}}=6$ on various datasets published in \cite{vanschoren2014openml}. An MTOP instance is constructed by randomly sampling two tasks in the task set, the same way as the BBOB problem set does. Then three HPO problem sets named `SVM', `XGBoost', and `FCNet' are formulated by constructing 100 MTOP instances based on each task set, respectively. We learn the agent from scratch on the three problem sets and test on unseen MTOP instances. The optimization horizon of learning and testing is set as $G_{\rm{roll}}=G_{\max}=100$. To fulfill the potential of the L2T, we conduct a grid search on the parameter space $b_2\in\{0.01,0.1,1,1,10,100\}$ to set up the proper $b_2$. It turns out $b_2=0.01$ works well for MTDE-L2T and $b_2=10$ works well for MTGA-L2T. Moreover, the target accuracy is set to $\xi_{\rm{SVM}}=1,\xi_{\rm{XGBoost}}=0.1,\xi_{\rm{FCNet}}=0.1$ for SVM, XGBoost, and FCNet, respectively. The results of MTDE-L2T and MTGA-L2T with the best parameter setting of $b_2$ compared with peer EMT algorithms on HPO problems are shown in Table S.XII in the supplementary material \cite{wu2024learning}. It can be observed that the proposed L2T-based EMT algorithms offer competitive performance, demonstrating the effectiveness of the proposed L2T framework in solving real-world application MTOPs.

\section{Conclusion}
This study presents the L2T framework to refine the KT process's adaptability, which leverages RL to autonomously adapt KT intensity and strategies for the effective transfer of knowledge across tasks along the EMT process. This approach marks a departure from the conventional, expert-driven KT, significantly decreasing the need for specialized labor. By employing parameterized neural networks to model the agent for performing KT, the L2T framework empowers an automatic discovery of effective KT policies for the target MTOPs at hand. Our experimental study, encompassing both synthetic and real-world scenarios, confirms that he L2T framework not only improves the flexibility of the KT process but also broadens the implicit EMT algorithm's capacity to handle diverse and intricate MTOPs with increased efficiency.

Looking ahead, we recognize the challenge posed by the limited number of real-world problem instances and plan to explore instance generation \cite{tang2021few} as a potential solution. Furthermore, it is worthy of exploring the representation learning approach to achieve end-to-end state feature extraction besides manual feature extraction.

\bibliographystyle{IEEEtran}
\bibliography{resources/L2T}

\begin{thebibliography}{10}
\providecommand{\url}[1]{#1}
\csname url@samestyle\endcsname
\providecommand{\newblock}{\relax}
\providecommand{\bibinfo}[2]{#2}
\providecommand{\BIBentrySTDinterwordspacing}{\spaceskip=0pt\relax}
\providecommand{\BIBentryALTinterwordstretchfactor}{4}
\providecommand{\BIBentryALTinterwordspacing}{\spaceskip=\fontdimen2\font plus
\BIBentryALTinterwordstretchfactor\fontdimen3\font minus
  \fontdimen4\font\relax}
\providecommand{\BIBforeignlanguage}[2]{{%
\expandafter\ifx\csname l@#1\endcsname\relax
\typeout{** WARNING: IEEEtran.bst: No hyphenation pattern has been}%
\typeout{** loaded for the language `#1'. Using the pattern for}%
\typeout{** the default language instead.}%
\else
\language=\csname l@#1\endcsname
\fi
#2}}
\providecommand{\BIBdecl}{\relax}
\BIBdecl

\bibitem{zhan2022survey}
Z.-H. Zhan, L.~Shi, K.~C. Tan, and J.~Zhang, ``A survey on evolutionary
  computation for complex continuous optimization,'' \emph{Artificial
  Intelligence Review}, pp. 1--52, 2022.

\bibitem{gupta2015multifactorial}
A.~Gupta, Y.-S. Ong, and L.~Feng, ``Multifactorial evolution: toward
  evolutionary multitasking,'' \emph{IEEE Transactions on Evolutionary
  Computation}, vol.~20, no.~3, pp. 343--357, 2015.

\bibitem{tan2021evolutionary}
K.~C. Tan, L.~Feng, and M.~Jiang, ``Evolutionary transfer optimization-a new
  frontier in evolutionary computation research,'' \emph{IEEE Computational
  Intelligence Magazine}, vol.~16, no.~1, pp. 22--33, 2021.

\bibitem{wu2023transferable}
S.-H. Wu, Z.-H. Zhan, K.~C. Tan, and J.~Zhang, ``Transferable adaptive
  differential evolution for many-task optimization,'' \emph{IEEE Transactions
  on Cybernetics}, vol.~53, no.~11, pp. 7295--7308, 2023.

\bibitem{feng2018evolutionary}
L.~Feng, L.~Zhou, J.~Zhong, A.~Gupta, Y.-S. Ong, K.-C. Tan, and A.~K. Qin,
  ``Evolutionary multitasking via explicit autoencoding,'' \emph{IEEE
  Transactions on Cybernetics}, vol.~49, no.~9, pp. 3457--3470, 2018.

\bibitem{chen2020evolutionary}
K.~Chen, B.~Xue, M.~Zhang, and F.~Zhou, ``An evolutionary multitasking-based
  feature selection method for high-dimensional classification,'' \emph{IEEE
  Transactions on Cybernetics}, vol.~52, no.~7, pp. 7172--7186, 2020.

\bibitem{feng2020explicit}
L.~Feng, Y.~Huang, L.~Zhou, J.~Zhong, A.~Gupta, K.~Tang, and K.~C. Tan,
  ``Explicit evolutionary multitasking for combinatorial optimization: A case
  study on capacitated vehicle routing problem,'' \emph{IEEE Transactions on
  Cybernetics}, vol.~51, no.~6, pp. 3143--3156, 2020.

\bibitem{chen2019adaptive}
Y.~Chen, J.~Zhong, L.~Feng, and J.~Zhang, ``An adaptive archive-based
  evolutionary framework for many-task optimization,'' \emph{IEEE Transactions
  on Emerging Topics in Computational Intelligence}, vol.~4, no.~3, pp.
  369--384, 2019.

\bibitem{zhou2020toward}
L.~Zhou, L.~Feng, K.~C. Tan, J.~Zhong, Z.~Zhu, K.~Liu, and C.~Chen, ``Toward
  adaptive knowledge transfer in multifactorial evolutionary computation,''
  \emph{IEEE Transactions on Cybernetics}, vol.~51, no.~5, pp. 2563--2576,
  2020.

\bibitem{ma2024metabox}
Z.~Ma, H.~Guo, J.~Chen, Z.~Li, G.~Peng, Y.-J. Gong, Y.~Ma, and Z.~Cao,
  ``Metabox: A benchmark platform for meta-black-box optimization with
  reinforcement learning,'' \emph{Advances in Neural Information Processing
  Systems}, vol.~36, 2024.

\bibitem{schulman2017proximal}
J.~Schulman, F.~Wolski, P.~Dhariwal, A.~Radford, and O.~Klimov, ``Proximal
  policy optimization algorithms,'' \emph{arXiv preprint arXiv:1707.06347},
  2017.

\bibitem{li2021meta}
J.-Y. Li, Z.-H. Zhan, K.~C. Tan, and J.~Zhang, ``A meta-knowledge
  transfer-based differential evolution for multitask optimization,''
  \emph{IEEE Transactions on Evolutionary Computation}, vol.~26, no.~4, pp.
  719--734, 2021.

\bibitem{gupta2017insights}
A.~Gupta, Y.-S. Ong, and L.~Feng, ``Insights on transfer optimization: Because
  experience is the best teacher,'' \emph{IEEE Transactions on Emerging Topics
  in Computational Intelligence}, vol.~2, no.~1, pp. 51--64, 2017.

\bibitem{bali2019multifactorial}
K.~K. Bali, Y.-S. Ong, A.~Gupta, and P.~S. Tan, ``Multifactorial evolutionary
  algorithm with online transfer parameter estimation: Mfea-ii,'' \emph{IEEE
  Transactions on Evolutionary Computation}, vol.~24, no.~1, pp. 69--83, 2019.

\bibitem{lin2023ensemble}
W.~Lin, Q.~Lin, L.~Feng, and K.~C. Tan, ``Ensemble of domain adaptation-based
  knowledge transfer for evolutionary multitasking,'' \emph{IEEE Transactions
  on Evolutionary Computation}, 2023.

\bibitem{wu2024learning}
S.-H. Wu, Y.~Huang, X.~Wu, L.~Feng, Z.-H. Zhan, and K.~C. Tan, ``Supplementary
  material for `learning to transfer for evolutionary multitasking','' 2024,
  [Online]. Available:
  \url{https://zhanapollo.github.io/zhanzhh/resources.htm}.

\bibitem{xu2021evolutionary}
H.~Xu, A.~K. Qin, and S.~Xia, ``Evolutionary multitask optimization with
  adaptive knowledge transfer,'' \emph{IEEE Transactions on Evolutionary
  Computation}, vol.~26, no.~2, pp. 290--303, 2021.

\bibitem{feng2017empirical}
L.~Feng, W.~Zhou, L.~Zhou, S.~Jiang, J.~Zhong, B.~Da, Z.~Zhu, and Y.~Wang, ``An
  empirical study of multifactorial pso and multifactorial de,'' in \emph{Proc.
  IEEE Congress on evolutionary computation (CEC)}, 2017, pp. 921--928.

\bibitem{jin2019study}
C.~Jin, P.-W. Tsai, and A.~K. Qin, ``A study on knowledge reuse strategies in
  multitasking differential evolution,'' in \emph{Proc. IEEE Congress on
  Evolutionary Computation (CEC)}, 2019, pp. 1564--1571.

\bibitem{wang2021solving}
C.~Wang, J.~Liu, K.~Wu, and Z.~Wu, ``Solving multitask optimization problems
  with adaptive knowledge transfer via anomaly detection,'' \emph{IEEE
  Transactions on Evolutionary Computation}, vol.~26, no.~2, pp. 304--318,
  2021.

\bibitem{yi2022automated}
W.~Yi, R.~Qu, L.~Jiao, and B.~Niu, ``Automated design of metaheuristics using
  reinforcement learning within a novel general search framework,'' \emph{IEEE
  Transactions on Evolutionary Computation}, 2022.

\bibitem{sun2021learning}
J.~Sun, X.~Liu, T.~B{\"a}ck, and Z.~Xu, ``Learning adaptive differential
  evolution algorithm from optimization experiences by policy gradient,''
  \emph{IEEE Transactions on Evolutionary Computation}, vol.~25, no.~4, pp.
  666--680, 2021.

\bibitem{zhan2022learning}
Z.-H. Zhan, J.-Y. Li, S.~Kwong, and J.~Zhang, ``Learning-aided evolution for
  optimization,'' \emph{IEEE Transactions on Evolutionary Computation},
  vol.~27, no.~6, pp. 1794--1808, 2023.

\bibitem{jiang2023knowledge}
Y.~Jiang, Z.-H. Zhan, K.~C. Tan, and J.~Zhang, ``Knowledge learning for
  evolutionary computation,'' \emph{IEEE Transactions on Evolutionary
  Computation}, 2023.

\bibitem{zhang2021learning}
H.~Zhang, J.~Sun, and Z.~Xu, ``Learning to mutate for differential evolution,''
  in \emph{Proc. IEEE Congress on Evolutionary Computation (CEC)}, 2021, pp.
  1--8.

\bibitem{khadka2018evolution}
S.~Khadka and K.~Tumer, ``Evolution-guided policy gradient in reinforcement
  learning,'' \emph{Advances in Neural Information Processing Systems},
  vol.~31, 2018.

\bibitem{da2017evolutionary}
B.~Da, Y.-S. Ong, L.~Feng, A.~K. Qin, A.~Gupta, Z.~Zhu, C.-K. Ting, K.~Tang,
  and X.~Yao, ``Evolutionary multitasking for single-objective continuous
  optimization: Benchmark problems, performance metric, and baseline results,''
  \emph{arXiv preprint arXiv:1706.03470}, 2017.

\bibitem{hansen2021coco}
N.~Hansen, A.~Auger, R.~Ros, O.~Mersmann, T.~Tu{\v{s}}ar, and D.~Brockhoff,
  ``Coco: A platform for comparing continuous optimizers in a black-box
  setting,'' \emph{Optimization Methods and Software}, vol.~36, no.~1, pp.
  114--144, 2021.

\bibitem{ding2017generalized}
J.~Ding, C.~Yang, Y.~Jin, and T.~Chai, ``Generalized multitasking for
  evolutionary optimization of expensive problems,'' \emph{IEEE Transactions on
  Evolutionary Computation}, vol.~23, no.~1, pp. 44--58, 2017.

\bibitem{xue2020affine}
X.~Xue, K.~Zhang, K.~C. Tan, L.~Feng, J.~Wang, G.~Chen, X.~Zhao, L.~Zhang, and
  J.~Yao, ``Affine transformation-enhanced multifactorial optimization for
  heterogeneous problems,'' \emph{IEEE Transactions on Cybernetics}, vol.~52,
  no.~7, pp. 6217--6231, 2020.

\bibitem{klein2019meta}
A.~Klein, Z.~Dai, F.~Hutter, N.~Lawrence, and J.~Gonzalez, ``Meta-surrogate
  benchmarking for hyperparameter optimization,'' \emph{Advances in Neural
  Information Processing Systems}, vol.~32, 2019.

\bibitem{vanschoren2014openml}
J.~Vanschoren, J.~N. Van~Rijn, B.~Bischl, and L.~Torgo, ``Openml: networked
  science in machine learning,'' \emph{ACM SIGKDD Explorations Newsletter},
  vol.~15, no.~2, pp. 49--60, 2014.

\bibitem{tang2021few}
K.~Tang, S.~Liu, P.~Yang, and X.~Yao, ``Few-shots parallel algorithm portfolio
  construction via co-evolution,'' \emph{IEEE Transactions on Evolutionary
  Computation}, vol.~25, no.~3, pp. 595--607, 2021.

\end{thebibliography}

\end{document}